%% file: iclr2027_conference.tex
\documentclass{article} 
\usepackage{iclr2027_conference,times}

\input{math_commands.tex}

\usepackage{hyperref}
\usepackage{url}

\usepackage{booktabs}
\usepackage{graphicx}
\usepackage{subcaption}
\usepackage{multirow}

\title{Query-adaptive Recurrent Tokenization}

\author{Bavesh Balaji, Pranav Gopinath, Pouya Sadeghi, Sirisha Rambhatla \\
Critical ML Lab\\
University of Waterloo\\
\texttt{\{bbalaji, p3gopina, pouya.sadeghi, sirisha.rambhatla\}@uwaterloo.ca} \\
}

\usepackage{xcolor}
\usepackage[most]{tcolorbox}
\usepackage{svg}
\usepackage{pifont}

\newcommand{\cmark}{\ding{51}}
\newcommand{\xmark}{\ding{55}}

\definecolor{insightbg}{RGB}{242,247,252}
\definecolor{insightborder}{RGB}{160,185,210}

\newtcolorbox{insightbox}{
    colback=insightbg,
    colframe=insightborder,
    boxrule=0.4pt,
    arc=1.5mm,
    left=1mm,
    right=1mm,
    top=0.8mm,
    bottom=0.8mm,
    enhanced,
    borderline west={1.2pt}{0pt}{insightborder},
    title=Insight,
    fonttitle=\bfseries,
    coltitle=black
}

\iclrfinalcopy 
\begin{document}

\maketitle
\pagestyle{plain}
\thispagestyle{plain}


\input{sections/abstract}
\input{sections/introduction}

\input{sections/related_work}

\input{sections/method}
\input{sections/experiments}
\input{sections/conclusion}
\input{sections/ai_use_statement}
\input{sections/ethics_statement}

\input{sections/reproducibility_statement}

\bibliography{iclr2027_conference}
\bibliographystyle{iclr2027_conference}

\appendix
\input{sections/appendix}

\end{document}

%% file: math_commands.tex
\usepackage{amsmath,amsfonts,bm}

\def\eqref#1{equation~\ref{#1}}

\def\1{\bm{1}}

\DeclareMathAlphabet{\mathsfit}{\encodingdefault}{\sfdefault}{m}{sl}
\SetMathAlphabet{\mathsfit}{bold}{\encodingdefault}{\sfdefault}{bx}{n}



%% file: sections/abstract.tex
\begin{abstract}

Visual information worth extracting from an image and the token budget it warrants depend on the text query. Yet modern vision and vision-language models largely encode images independently of the query and allocate a fixed visual token budget to every input. We introduce the Query-Adaptive Recurrent Tokenizer (QART), a backbone-agnostic module that makes visual representations both query-\textit{steerable} and budget-\textit{adaptive}. QART steers the visual representation toward query-relevant evidence while adapting its token budget to the amount of information required. Its key principle is simple. Additional tokens are allocated until the visual features relevant to the query are represented with sufficient fidelity. Starting from frozen encoder features, QART recurrently constructs a compact representation and uses reconstruction quality over the query-relevant visual support to control token allocation. On Visual7W and VQAv2 , QART achieves \textbf{81.96\%} and \textbf{76.37\%} accuracy using only about 29 visual tokens, reducing token counts by \textbf{20–25$\times$} relative to dense VLM inputs. It further reaches 92.0\% P@1 on CORE and 36.3\% PR-AUC on PODS, outperforming SteerViT on PODS by 8.4\%. Together, these results show that visual representations need not be fixed in either content or capacity and can instead adapt to what each query requires.

\end{abstract}

%% file: sections/introduction.tex
\section{Introduction}
\label{sec:introduction}

\begin{quote}
\textit{``A wealth of information creates a poverty of attention and a need to allocate that attention efficiently among the overabundance of information sources that might consume it.''}

\raggedleft
--- Herbert A. Simon (\citeyear{simon1971designing})
\end{quote}

\begin{figure}[t]
    \centering
    \includegraphics[width=\textwidth]{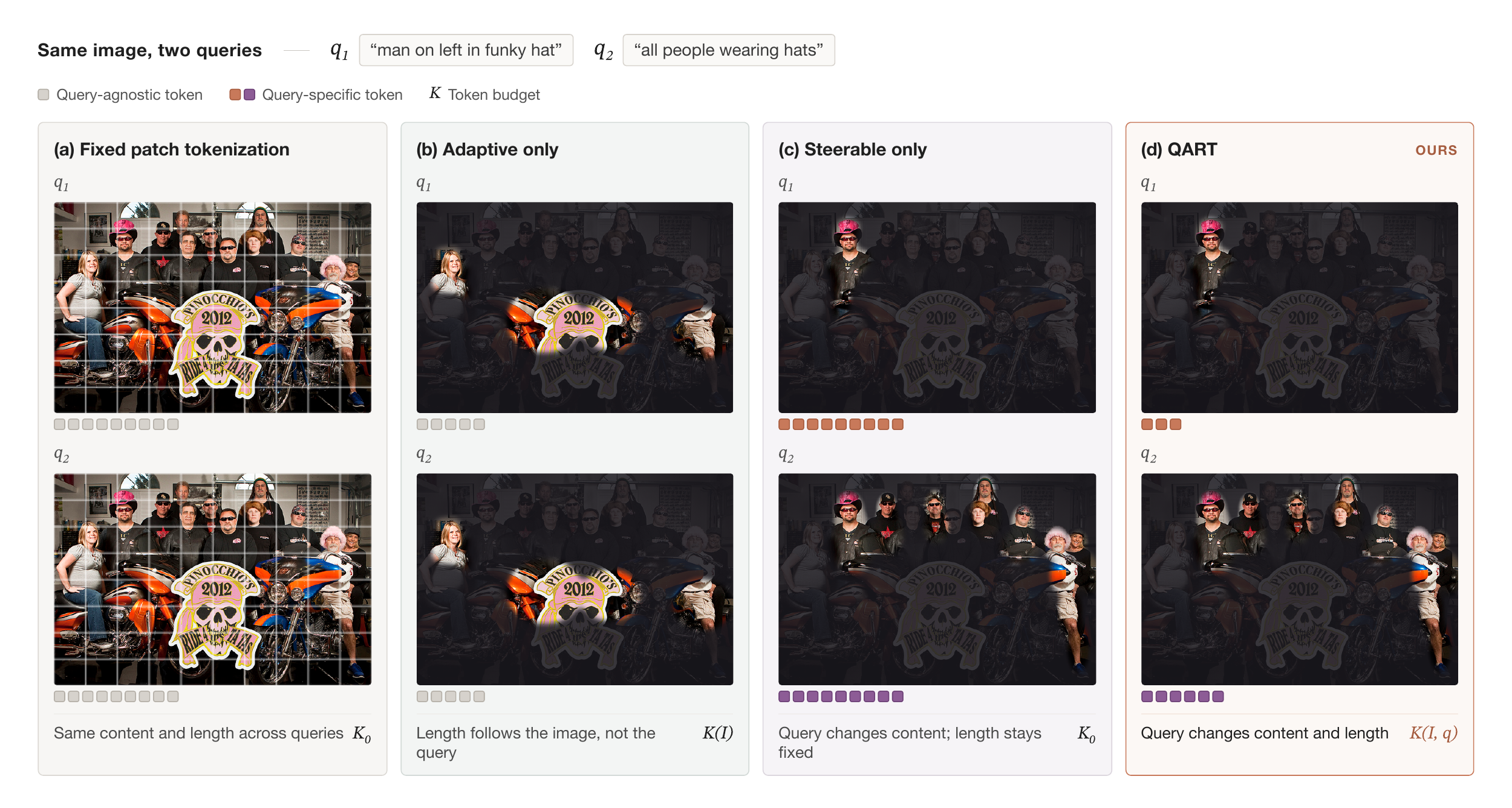}
    \caption{\textbf{QART combines query steerability with budget adaptivity.}
Fixed patch tokenizers use the same representation and token budget for every query. Adaptive tokenizers change \emph{how many} tokens are allocated, while steerable methods change \emph{what} visual information is emphasized. QART does both, steering the representation toward the query while adapting its capacity to the required visual evidence. The clear regions represent the tokenizers' focus.}
    \label{fig:teaser}
    \vspace{-1em}
\end{figure}

An image contains evidence for many possible questions, and the visual information that matters depends on what is being asked \cite{anderson2018bottomup, hudson2019gqa, pantazopoulos2025visualgrounding}. In the same street scene, identifying the color of a traffic light requires evidence from a small region, whereas counting vehicles turning left requires information distributed across objects and their spatial relationships \citep{acharya2018tallyqa, rajabi2024gsrbench, ishihara2025strideqa, huang2026evidencerlevidenceintensivevisualreasoning, pinpoint}. Yet SOTA vision encoders \citep{tschannen2025siglip2, radford2021learning, oquab2023dinov2, he2022masked} produce the same representation of an image regardless of the query (Figure ~\ref{fig:teaser}(a)). These representations capture broadly useful visual information but cannot explicitly prioritize the requested evidence \citep{ruthardt2026steervit}. More importantly, visual salience is not akin to query relevance: a prominent object may be irrelevant to the question, while a small detail may determine the answer. A useful visual representation should therefore be \textit{steerable}: changing the query should change which visual information the representation emphasizes.

A separate limitation concerns how much representational capacity is allocated to that evidence. Visual tokenizers commonly produce a fixed-size sequence; ViT-based encoders \citep{oquab2023dinov2, tschannen2025siglip2} provides a fixed set of patch tokens,  while the Q-Former uses a fixed number of latent tokens regardless of the image or the evidence required by the downstream task \citep{li2023blip2}. A large token budget increases downstream computation for every example \citep{yin2025dart, yin2022avit, li2026occamtoken}, while a small budget may discard necessary details \citep{kim2026and}. These considerations expose two distinct axes of visual tokenization: which evidence to preserve and how much capacity to allocate (\textit{adaptivity}), as shown in Fig.~\ref{fig:teaser}.
These properties are increasingly relevant to embodied systems, where agents repeatedly interpret visual observations under changing goals, and efficient processing matters over long interaction horizons \citep{saxena2025grapheqa,cheng2024efficienteqa}.
We seek a tokenizer that connects both axes, allocating capacity according to how well the current representation preserves query-relevant visual evidence.

Prior work addresses parts of this problem. Learned resamplers create compact latent representations, steerable encoders condition visual features on language \citep{ruthardt2026steervit}, and token-pruning methods remove redundant or query-irrelevant patches \citep{li2026occamtoken, takezoe2026learnpruner}. Adaptive image tokenizers additionally vary their sequence length according to image complexity \citep{duggal2024alit, yu2024image}. What remains less explored is an explicit mechanism that couples these choices: rather than prescribing a token count or predicting one directly, can a tokenizer continue allocating capacity until the evidence relevant to the current query has been represented with sufficient fidelity?

We introduce the Query-Adaptive Recurrent Tokenizer (QART), a visual tokenizer that converts dense encoder features into a variable-length set of query-conditioned visual tokens. Given frozen image features and a text query, a cross-modal relevance module first identifies the visual evidence relevant to the query. QART begins with a small set of tokens and recurrently refines them using this evidence and the query. After each round, a shared decoder reconstructs the relevant visual features, and reconstruction quality determines whether more tokens are added or computation halts. The resulting tokens form a compact representation whose content and capacity are both shaped by the query, and can be used by a downstream task head or language model.


We evaluate QART on prompt-conditioned retrieval (CORE \citep{ruthardt2026steervit}, GeneCIS \citep{vaze2023genecis}, PODS \citep{sundaram2025personalized}), compositional matching (Winoground \citep{thrush2022winoground}), Visual Question Answering (VQAv2 \citep{goyal2017vqav2}), and grounding (Visual7W pointing \citep{zhu2016cvpr}. QART learns non-uniform allocations and compresses dense patch tokens by ~25$\times$. On CORE and GeneCIS, it improves over the query-agnostic SigLIP2 representation by 23.5\% and 2.7\% respectively. On Visual7W pointing, it outperforms the strongest token-reduction baseline DART \cite{wen2025stop} by 22.1\% under a matched protocol; on VQAv2, it retains 93.8\% of dense accuracy while reducing tokens by 25$\times$ and outperforms a matched 32-token Q-Former, while improving end-to-end throughput by 1.53$\times$. Together, these results highlight the significance of query-adaptive visual representations.



%% file: sections/related_work.tex
\section{Related Work}
\label{sec:related_work}

\subsection{Compact and Variable-Length Visual Tokenization}

Vision encoders commonly represent an image as a dense grid of patch embeddings, motivating learned bottlenecks that summarize this grid with a smaller set of visual tokens \citep{bachmann2025flextok}. TokenLearner learns a compact set of content-adaptive tokens \citep{ryoo2021tokenlearner}, while BLIP-2 uses a Querying Transformer (Q-Former) to map frozen image features into a fixed-size latent sequence for a downstream language model \citep{li2023blip2}. InstructBLIP further conditions the Q-Former on the input instruction, but retains a fixed latent budget \citep{dai2023instructblip}. Matryoshka Query Transformer (MQT) supports multiple output lengths within one model by training nested prefixes of latent queries; however, the desired length is selected externally at inference time rather than inferred from the visual evidence required by each query \citep{hu2024mqt}.

Adaptive-length image tokenizers vary representation capacity with image content \citep{yan2024elastictok,duggal2025karl,chen2026stat,lu2026adatok}. ALIT, the closest architectural precedent to our work, recurrently updates a latent sequence, adds tokens over successive rounds, and uses reconstruction as a learning signal for variable-length image representations \citep{duggal2024alit}. QART changes the unit of adaptation from the image alone to the image--query pair. Its text-conditioned relevance model determines which patch features must be preserved, and the recurrent tokenizer halts when those query-relevant features have been reconstructed sufficiently well. Consequently, the same image can produce different token contents and lengths under different queries.

\subsection{Query-Conditioned and Steerable Visual Representations}

Other work studies how language can steer visual representation \citep{mao2025dove}. InstructBLIP extracts instruction-specific features through an instruction-aware Q-Former \citep{dai2023instructblip}; FLAIR applies text-conditioned attention pooling to local image features for fine-grained retrieval \citep{xiao2024flair}; and TIE conditions the image encoder on the text query \citep{thirukovalluru2025tie}. SteerViT injects language into intermediate vision-transformer layers to produce dense visual features that can be directed toward a requested concept \citep{ruthardt2026steervit}. QMoP uses the query to route visual features among pooling, resampling, and pruning branches \citep{li2026qmop}.

These methods establish that useful visual evidence depends on the language context; QART is therefore not the first query-conditioned visual representation. Its distinction lies in coupling query conditioning to recurrent capacity allocation. Rather than only changing which information is emphasized within a dense or prescribed-size representation, QART jointly determines what information to encode and how many latent tokens are required. The token count follows from an explicit representation-adequacy criterion instead of a fixed budget or a separately predicted length.

\subsection{Visual Token Pruning and Merging}

Visual-token reduction is also widely studied through pruning and merging \citep{chen2024fastv,zhang2025sparsevlm,yang2025visionzip,zhang2025vispruner,shang2025prumerge}. ToMe reduces vision-transformer sequences by progressively merging similar tokens \citep{bolya2023tome}. In vision-language systems, QG-VTC uses question--patch relevance to progressively compress visual tokens within the vision encoder \citep{li2025qgvtc}, while QuietPrune introduces query guidance early in the vision transformer so that pruning preserves prompt-relevant regions \citep{gao2026quietprune}. ATP-LLaVA learns instance- and layer-dependent pruning ratios inside the language model \citep{ye2024atpllava}; LearnPruner combines learned post-encoder reduction with query-guided pruning in intermediate language-model layers \citep{takezoe2026learnpruner}; and OccamToken derives image- and query-adaptive pruning thresholds without additional training \citep{li2026occamtoken}.

These approaches primarily retain, merge, or discard existing tokens. QART instead constructs and recurrently expands a slot-based representation from frozen patch features until a query-conditioned reconstruction criterion is met, which supplies a shorter visual sequence to downstream models.

%% file: sections/method.tex
\section{Method}
\label{sec:method}

\begin{figure}[t]
    \centering
    \includegraphics[width=\textwidth]{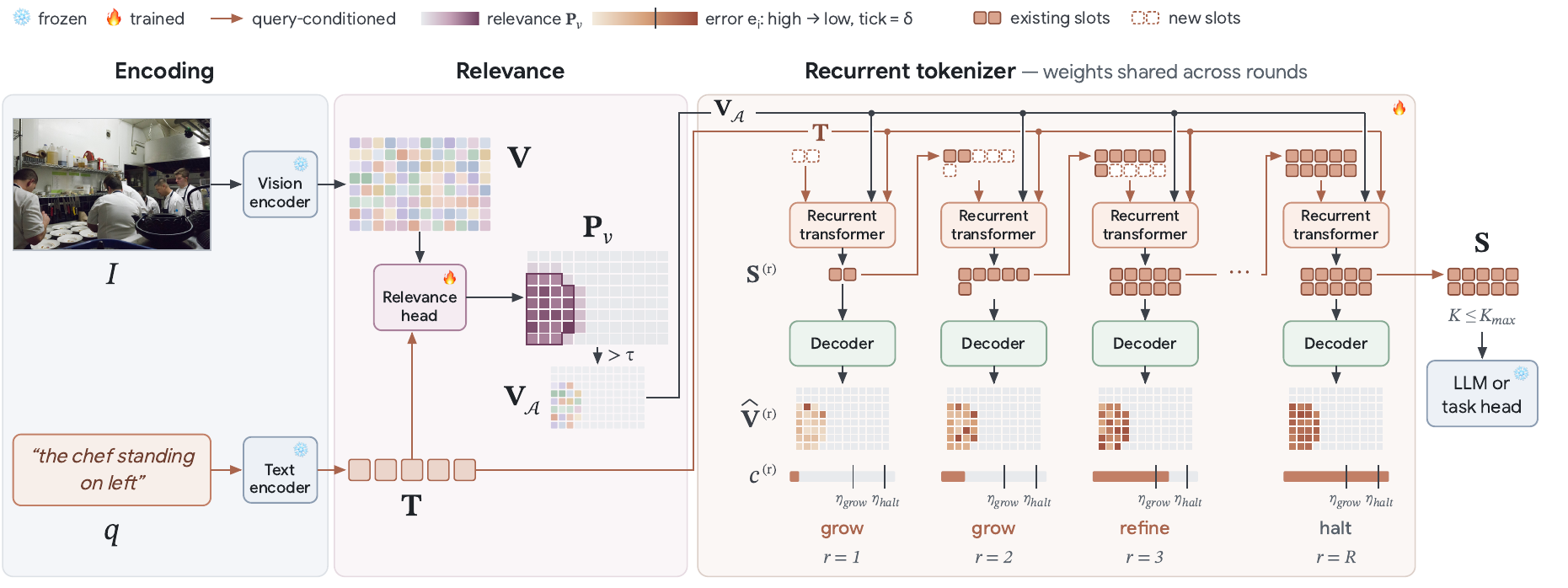}
    \caption{
        \textbf{Overview of QART.}
        Given an image and a text query, frozen encoders provide visual and
        linguistic features $\mathbf{V}$ and $\mathbf{T}$ respectively. A relevance module $H_{\phi}$ selects query-relevant patches $\mathbf{V}_{\mathcal{A}}$;
        QART then recurrently refines a small set of query-conditioned slots $\mathbf{S}$
        by passing them along with $\mathbf{V}_{\mathcal{A}}$ and $\mathbf{T}$ to the shared transformer $F_{\theta}$. After each round $r$, a shared decoder $D_{\psi}$
        reconstructs the visual features, and reconstruction quality
        determines whether more slots are added or recurrence halts.
        Only slot states propagate across rounds.
    }
    \label{fig:methodology}
\end{figure}

QART converts dense encoder features into a variable-length sequence of query-conditioned visual tokens in three stages (Figure~\ref{fig:methodology}).
A cross-modal relevance module selects query-relevant visual features (\S\ref{sec:relevance}); a weight-shared recurrent Transformer then builds a slot-based representation from them, beginning with $K_0$ query-initialized slots and adding $\Delta K$ new ones when reconstruction coverage of the selected features falls below a coverage criterion (\S\ref{sec:recurrent_tokenizer}--\ref{sec:halting}).
The trainable components are optimized jointly using query-weighted
reconstruction and relevance-prediction objectives (\S\ref{sec:objective}).

\subsection{Identifying Query-Relevant Evidence}
\label{sec:relevance}

Given an image $I \in \mathbb{R}^{H \times W \times 3}$ and text query~$q$, QART represents the image--query pair as a variable-length set of steered visual tokens, or \emph{slots}:
\begin{equation}
    G_{\theta, \phi, \psi}(I, q) = \mathbf{S} \in \mathbb{R}^{K \times D},
    \qquad K \leq K_{\max}.
    \label{eq:qart_output}
\end{equation}
Frozen vision and text encoders produce visual features $\mathbf{V} = E_{\mathrm{v}}(I) \in \mathbb{R}^{N \times D_v}$ and text tokens $\mathbf{T} = E_{\mathrm{t}}(q) \in \mathbb{R}^{L \times D_t}$, projected to a shared dimension~$D$.
A cross-modal relevance module then predicts a relevance map $\mathbf{P}_\mathrm{v}$ which indicates how strongly each patch~$\mathbf{v}_i$ corresponds to the query:
\begin{equation}
    \mathbf{P}_{\mathrm{v}} = [p_1, p_2, ..., p_N]^T \in [0,1]^N, \qquad p_i = H_{\phi}(\mathbf{v}_i, \mathbf{T}),
    \qquad p_i \in [0, 1].
    \label{eq:relevance}
\end{equation}
This relevance map is used to identify the query-relevant support $\mathcal{A} = \{i \mid p_i > \tau\}$ and the corresponding features $\mathbf{V}_{\mathcal{A}} = \{\mathbf{v}_i\}_{i \in \mathcal{A}}$.
Having identified which evidence to preserve, the remaining problem is how to represent~$\mathbf{V}_{\mathcal{A}}$ compactly without prescribing a fixed token count.

\subsection{Recurrent Query-Conditioned Representation}
\label{sec:recurrent_tokenizer}

To answer the question, QART builds on progressive token allocation~\citep{duggal2024alit}. Specifically, a pooled query representation~$\bar{\mathbf{q}}$ seeds $K_0$ initial slots through a learned projector, $\mathbf{S}^{(0)} = f_{\mathrm{init}}(\bar{\mathbf{q}}) \in \mathbb{R}^{K_0 \times D}$;these slots are then recurrently refined to gather evidence from~$\mathbf{V}_{\mathcal{A}}$.

At each round~$r$, the slots $\mathbf{S}^{(r-1)}$ interact with the selected visual features $\mathbf{V}_{\mathcal{A}}$ and text through a shared Transformer~$F_{\theta}$, followed by a cross-attention readout over~$\mathbf{V}_{\mathcal{A}}$:
\begin{equation}
    \widetilde{\mathbf{S}}^{(r)}
    = F_{\theta}\!\left(\mathbf{S}^{(r-1)},\, \mathbf{V}_{\mathcal{A}},\, \mathbf{T}\right),
    \qquad
    \mathbf{S}^{(r)}
    = \mathrm{LN}\!\left(
        \widetilde{\mathbf{S}}^{(r)}
        + \mathrm{MHA}\!\left(
            \widetilde{\mathbf{S}}^{(r)},\, \mathbf{V}_{\mathcal{A}},\, \mathbf{V}_{\mathcal{A}}
        \right)
    \right).
    \label{eq:recurrent}
\end{equation}
%
When the current set of slots cannot represent $\mathbf{V}_{\mathcal{A}}$ with sufficient fidelity , $\Delta K$ new slots are initialized from~$\bar{\mathbf{q}}$ via $f_{\mathrm{init}}$ and concatenated to the existing set before the next round, growing the representation from $K_r$ to $\min(K_r + \Delta K,\, K_{\max})$.

\subsection{Reconstruction-Guided Token Allocation}
\label{sec:halting}

QART ties representation sufficiency to feature reconstruction. Specifically, 
after each round, a shared decoder~$D_{\psi}$ reconstructs the full visual feature grid from the current slots, producing $\widehat{\mathbf{V}}^{(r)} = D_{\psi}(\mathbf{S}^{(r)})$.
The per-patch reconstruction error is $|\widehat{\mathbf{v}}_i^{(r)} - \mathbf{v}_i\|_2^2 / D$, evaluated only over~$\mathcal{A}$.
The fraction of support patches whose error has dropped below a threshold~$\delta$ by round~$r$ defines the \emph{reconstruction coverage}:
\begin{equation}
    c^{(r)}
    = \frac{1}{|\mathcal{A}|}
      \sum_{i \in \mathcal{A}}
      \mathbb{I}\!\left[\\|\widehat{\mathbf{v}}_i^{(r)} - \mathbf{v}_i\|_2^2 / D < \delta\right].
    \label{eq:coverage}
\end{equation}
If $c^{(r)} < \eta_{\mathrm{grow}}$, $\Delta K$ new slots are added before round $r{+}1$:
\begin{equation}
    K_{r+1}
    = \begin{cases}
        \min(K_r + \Delta K,\; K_{\max}), & c^{(r)} < \eta_{\mathrm{grow}}, \\[1mm]
        K_r, & \text{otherwise}.
    \end{cases}
    \label{eq:growth}
\end{equation}
Recurrence halts once $c^{(r)} \geq \eta_{\mathrm{halt}}$.
Setting $\eta_{\mathrm{grow}} < \eta_{\mathrm{halt}}$ decouples the two decisions: slot addition can stop while the existing slots continue to refine.
Because~$\mathcal{A}$ depends on the query, the same image follows different allocation trajectories under different queries (App.~\ref{sec:qualitative_adaptivity}).

\subsection{Learning Objective}
\label{sec:objective}

Given soft relevance targets $\mathbf{h} \in [0,1]^N$ derived from segmentation or alignment annotations (App.~\ref{app:datasets}), we define per-patch importance weights $\rho_i^{*} = h_i / (\sum_j h_j + \epsilon)$ and the target support $\mathcal{A}^{*} = \{i \mid h_i > 0\}$.
The reconstruction loss weights each patch by its query relevance, averaged across all~$R$ rounds:
\begin{equation}
    \mathcal{L}_{\mathrm{rec}}
    = \frac{1}{R}
      \sum_{r=1}^{R}
      \sum_{i \in \mathcal{A}^{*}}
      \rho_i^{*}\, |\widehat{\mathbf{v}}_i^{(r)} - \mathbf{v}_i\|_2^2 / D.
    \label{eq:loss_rec}
\end{equation}
The relevance module is trained with soft binary cross-entropy against~$\mathbf{h}$, denoted~$\mathcal{L}_{\mathrm{rel}}$.
The combined objective is
\begin{equation}
    \mathcal{L}
    = \lambda_{\mathrm{rec}}\,\mathcal{L}_{\mathrm{rec}}
    + \lambda_{\mathrm{rel}}\,\mathcal{L}_{\mathrm{rel}}.
    \label{eq:loss_total}
\end{equation}
At inference, no relevance supervision is needed; the predicted relevance map and reconstruction coverage jointly determine the token count and number of recurrent rounds for each image--query pair.

%% file: sections/experiments.tex
\section{Experiments}
\label{sec:experiments}

Our experiments probe the central properties of QART rather than performance on individual benchmarks alone. We ask whether query conditioning can steer a frozen visual representation toward the requested content, whether the resulting compact representation preserves downstream reasoning, whether its behavior genuinely depends on the query, whether it retains information beyond what the query explicitly specifies, and whether its compactness translates into practical inference gains.

\subsection{Experimental Setup}
\label{sec:implementation}

Unless stated otherwise, QART uses frozen SigLIP2 vision and text encoders and produces at most 32 visual tokens from 729 patch features. A single pretrained QART checkpoint is used across the representation benchmarks without task-specific finetuning. For VQA, QART replaces the visual tokens of LLaVA-1.5-7B while both QART and the language model remain frozen; only a lightweight visual projector is trained. Full architecture, training, and evaluation details are provided in the appendix \ref{app:implementation}. We additionally evaluate backbone generalization by using DINOv2 in App.~\ref{app:additional_results}.

\begin{figure*}[t]
    \centering

    \begin{subfigure}[t]{0.245\textwidth}
        \centering
        \includegraphics[width=\linewidth]{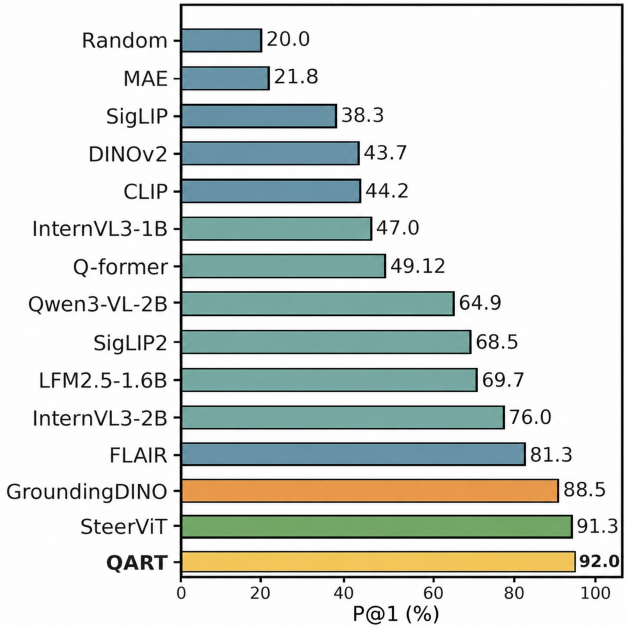}
        \caption{Steering on CORE.}
        \label{fig:rep_core}
    \end{subfigure}
    \hfill
    \begin{subfigure}[t]{0.245\textwidth}
        \centering
        \includegraphics[width=\linewidth]{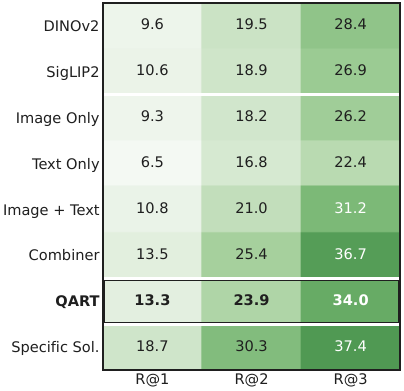}
        \caption{Zero-shot transfer on GeneCIS.}
        \label{fig:rep_genecis}
    \end{subfigure}
    \hfill
    \begin{subfigure}[t]{0.225\textwidth}
        \centering
        \includegraphics[width=\linewidth]{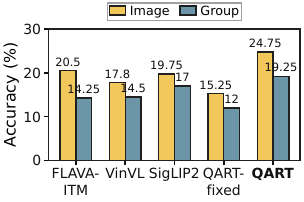}
        \caption{Query dependence on Winoground.}
        \label{fig:rep_winoground}
    \end{subfigure}
    \hfill
    \begin{subfigure}[t]{0.215\textwidth}
        \centering
        \includegraphics[width=\linewidth]{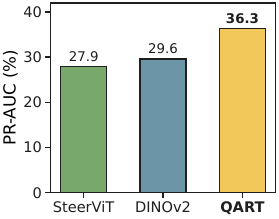}
        \caption{Fine-grained retention on PODS.}
        \label{fig:rep_pods}
    \end{subfigure}

    \caption{
        \textbf{QART steers visual representations while preserving transferable visual information.}
        (a) On CORE, QART redirects visual representations toward queried content.
        (b) The same representation transfers zero-shot to natural images on GeneCIS without dataset-specific tuning.
        (c) Replacing the sample-specific query with fixed conditioning substantially degrades Winoground performance.
        (d) On PODS, coarse conditioning preserves fine-grained instance information beyond the prompt.
    }
    \label{fig:representation_results}
\end{figure*}

\subsection{Does QART Redirect a Pretrained Visual Representation?}
\label{sec:results_retrieval}

We first ask whether QART changes what a frozen visual representation encodes. If query conditioning is effective, the resulting representation should emphasize the visual evidence requested by the query rather than simply inherit the dominant semantics of the pretrained backbone.

\paragraph{Controlled evaluation on CORE.}
CORE \cite{ruthardt2026steervit} \footnote{We follow SteerViT's protocol for CORE.} directly tests this behavior by requiring retrieval according to a queried object that may occupy only a small part of the image. QART reaches $92.0\%$ P@1, compared with $68.5\%$ for the native SigLIP2 global representation (Fig.~\ref{fig:rep_core}). 
Because both representations originate from the same frozen vision encoder, this improvement isolates the effect of query-conditioned representation construction. QART therefore does more than compress the underlying patch features; it changes which visual evidence dominates the representation. QART also outperforms open-vocabulary localization methods like Grounding-DINO \citep{liu2024grounding} and SteerViT \citep{ruthardt2026steervit} showcasing the benefit of query-relevant adaptive representations.


\paragraph{Does the same behavior transfer to natural images?}
We next test whether this steering behavior transfers beyond the controlled CORE setting. On GeneCIS Focus Object, QART uses the same pretrained checkpoint without GeneCIS-specific tuning. It improves over native SigLIP2 representations by $2.7$, $5.0$, and $7.1$\% at R@1, R@2, and R@3, respectively (Fig.~\ref{fig:rep_genecis}).

The gain increases with retrieval depth, suggesting that query conditioning consistently promotes condition-compatible images throughout the ranked list rather than affecting only the top prediction. QART also improves over simple Image + Text composition at every rank and approaches Combiner, despite Combiner being designed specifically for composed image-text retrieval.

\begin{insightbox}
\textbf{QART can steer a frozen visual representation toward query-relevant content, and this behavior transfers from controlled scenes to natural images without task-specific adaptation.}
\end{insightbox}

\begin{table*}[t]
    \centering

    \begin{subtable}[t]{0.49\textwidth}
        \centering
        \caption{VQAv2 results. Many baselines use a different vision tower,  so we use relative accuracies with the corresponding dense baselines to compare performance.}
        \label{tab:vqav2_results}
        \resizebox{\linewidth}{!}{
        \begin{tabular}{lcccc}
            \toprule
            Method & Base Model & Visual tokens $\downarrow$ & Accuracy $\uparrow$ & RelAcc\\
            \midrule
            LLaVA-1.5-7B & CLIP & 576 & 78.5 & 100\%
            \\
            LLaVA-1.5-7B+SigLIP2 & SigLIP2 & 729 & 81.4 & 100\% \\
            \midrule
            Q-Former & SigLIP2 & 32 & 74.67 & 91.73\%\\
            QG-VTC & CLIP & 36 & 71.4 & 90.96\%\\
            QMoP & CLIP & 36 & 74.1 & 94.39\%\\
            DivPrune      & CLIP & 32 & 71.2 & 90.7\% \\
            VisPruner     & CLIP & 32 & 67.7 & 86.24\% \\
            VisionZip     & CLIP & 32 & 67.1 & 85.48\% \\
            DART          & CLIP & 32 & 67.1 & 85.48\% \\
            FastV         & CLIP & 32 & 43.4 & 55.29\% \\
            SparseVLM     & CLIP & 32 & 58.6 & 74.65\% \\
            OccamToken    & CLIP & 32 & 75.5 & \textbf{96.18\%} \\
            LearnPruner   & CLIP & 32 & 74.0 & 94.27\% \\
            PruMerge+     & CLIP & 32 & 65.6 & 83.57\% \\
            \midrule
            \textbf{QART} & SigLIP2 & \textbf{29.4} & \textbf{76.37} & 93.82\% \\
            \bottomrule
        \end{tabular}
        }
    \end{subtable}
    \hfill
    \begin{subtable}[t]{0.48\textwidth}
        \centering
        \caption{Visual7W results. Here, all token-reduction methods are reproduced using the SigLIP2 vision tower so direct accuracy comparisons stand.}
        \label{tab:Visual7W_results}
        \resizebox{\linewidth}{!}{
        \begin{tabular}{lccc}
            \toprule
            Method & Visual tokens $\downarrow$ & Layer avg. $\downarrow$ & Accuracy $\uparrow$ \\
            \midrule
            LLaVA-1.5-7B & 576 & 576 & 83.21 \\
            LLaVA-1.5-7B+SigLIP2 & 729 & 729 & 88.49 \\
            \midrule
            DivPrune      & 32 & 32    & 52.16 \\
            VisPruner     & 32 & 32    & 49.35 \\
            VisionZip     & 32 & 32    & 46.82 \\
            DART          & 32 & 75.6  & 59.86 \\
            FastV         & 32 & 75.6  & 52.12 \\
            SparseVLM     & 32 & 118.0 & 51.92 \\
            OccamToken    & 32 & 129.1 & 55.61 \\
            LearnPruner   & 32 & 56.0  & 56.29 \\
            \midrule
            \textbf{QART} & \textbf{29} & \textbf{29} & \textbf{81.96} \\
            \bottomrule
        \end{tabular}
        }
    \end{subtable}

    \caption{
        \textbf{Reasoning and grounding under aggressive visual compression.}
        QART preserves strong downstream performance with approximately 29 visual tokens on both VQAv2 and Visual7W.
        Visual7W additionally reports the average token count processed across language-model layers for methods that prune within the LLM.
    }
    \label{tab:vlm_results}
\end{table*}

\subsection{Can QART Preserve Reasoning and Grounding Under Extreme Visual Compression?}
\label{sec:results_vlm}

Steering a representation is useful only if the resulting compact tokens retain the information required downstream. We test this in two complementary settings. VQAv2 evaluates general visual reasoning, while Visual7W places greater emphasis on grounding and localized visual evidence.

\paragraph{General reasoning on VQAv2.}
Replacing the dense visual tokens of LLaVA-1.5-7B+SigLIP2 with QART tokens, QART reaches 76.37\% acc., retaining 93.8\% of dense accuracy (81.4\%) with 24.8× fewer tokens, and outperforms a matched 32-token Q-Former by 1.7\%. Pruning baselines use a different base model, so we compare relative accuracies (Tab. \ref{tab:vqav2_results}); strong pruners retain slightly more performance here, as VQAv2's general questions limit the benefit of query-directed representation.

\paragraph{Grounded reasoning on Visual7W.}
We next test whether this compact representation preserves the spatially localized evidence required for grounding in Tab.~\ref{tab:Visual7W_results}. We specifically focus on the pointing task as it requires selecting the box that answers a question, so evidence at specific locations must be preserved. Under a matched pipeline (App. \ref{app:vis7w_eval}), QART reaches 81.96\% with 29 tokens, retaining 92.6\% of dense model's accuracy and outperforming the strongest baseline, DART, by 22.1\%. This larger gap indicates that aggressive compression is more demanding when predictions depend on localized visual evidence. Nevertheless, QART retains substantially more grounding information than existing low-token pruning methods. This suggests that query-conditioned token construction preserves the spatial evidence needed to associate the query with relevant image regions.

\begin{figure*}[t]
    \centering

    \begin{subfigure}[t]{0.36\textwidth}
        \centering
        \includegraphics[width=\linewidth]{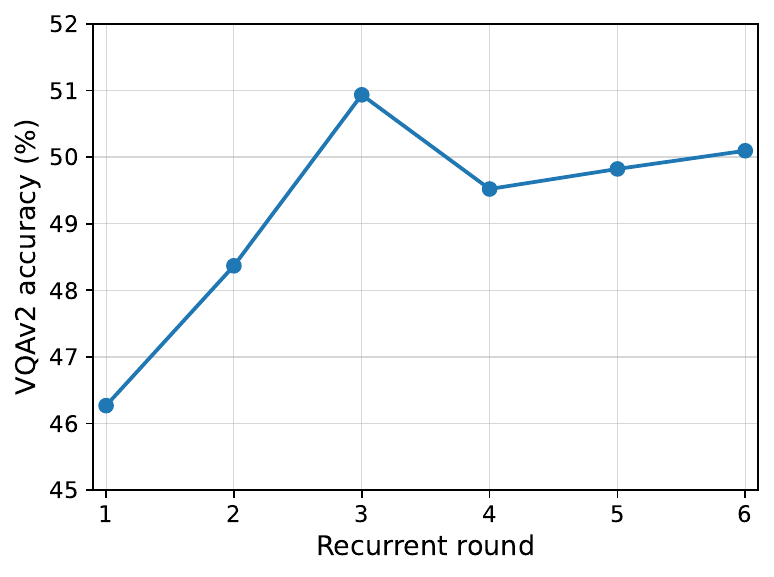}
        \caption{VQAv2 accuracy with a fixed $K=2$ token budget across rounds.}
        \label{fig:eff_budget}
    \end{subfigure}
    \hfill
    \begin{subfigure}[t]{0.3\textwidth}
        \centering
        \includegraphics[width=\linewidth]{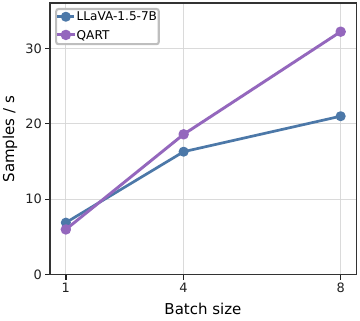}
        \caption{End-to-end VLM throughput across batch sizes.}
        \label{fig:eff_throughput}
    \end{subfigure}
    \hfill
    \begin{subfigure}[t]{0.3\textwidth}
        \centering
        \includegraphics[width=\linewidth]{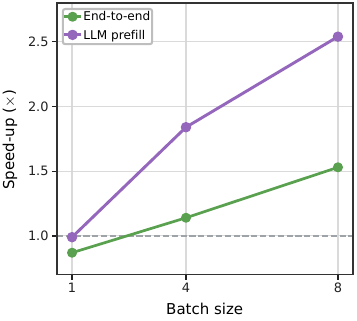}
        \caption{End-to-end and language-model prefill speed-up over LLaVA-1.5-7B.}
        \label{fig:eff_speedup}
    \end{subfigure}

    \caption{
        \textbf{Recurrent refinement and inference efficiency.}
        (a) QART's recurrence refines representations even upon disabling token growth.
        (b--c) The compact visual sequence increasingly improves throughput and language-model prefill as batch size grows.
    }
    \label{fig:efficiency_results}
\end{figure*}

\paragraph{Does recurrence refine the representation?}
To isolate recurrent refinement from token growth, we evaluate the same QART checkpoint while fixing the representation to two slots throughout recurrence, such that no additional slots are added. VQAv2 accuracy increases from $46.27\%$ after one round to $50.93\%$ after three rounds and remains around that number thereafter (Fig.~\ref{fig:eff_budget}). Since both the query-relevant support and token count remain fixed, this improvement shows that recurrent updates can refine the representation independently of increasing its capacity. Additional qualitative results are provided in App.~\ref{sec:recurrence_analysis}.

\begin{insightbox}
\textbf{QART preserves general reasoning and excels at grounding under severe compression.} With ~29 tokens, it retains over 92\% of dense tokens' accuracy on both benchmarks, with the largest gains when answers depend on localized evidence.
\end{insightbox}

\subsection{Does the Conditioning Query Actually Control the Representation?}
\label{sec:results_winoground}

We use Winoground \citep{thrush2022winoground} to test whether QART responds to the compositional meaning of the conditioning query. The benchmark pairs two images with two captions that contain the same words but differ in composition, and evaluates whether the model matches each caption to the correct image. We emphasize Image accuracy, where both candidate images are conditioned on the same caption, and Group accuracy, which additionally requires the Text judgment to be correct.

QART reaches $24.75\%$ Image and $19.25\%$ Group accuracy, compared with $19.75\%$ and $17.00\%$ for native SigLIP2. It also has 36.25\% text accuracy. Further, it outperforms stronger models like FLAVA \citep{singh2022flava} and Vin-VL \citep{hu2022scaling} comfortably. More importantly, replacing the sample-specific query with a fixed query reduces Image accuracy to $15.25\%$ and Group accuracy to $12.00\%$ (Fig.~\ref{fig:rep_winoground}).

This controlled intervention isolates the effect of conditioning from the tokenizer itself, showing that QART changes its visual representation according to the compositional meaning of the query rather than only the words it contains. Sec.~\ref{sec:query_conditioning_ablation}. further probes the impact of query
conditioning in representation steering and adaptive token allocation.

\subsection{Does Query Conditioning Preserve Information Beyond the Prompt?}
\label{sec:results_pods}

Query-conditioned compression introduces a complementary risk. A representation may become highly focused on the prompt while discarding visual detail that the prompt does not explicitly mention. We test this on PODS, where QART receives only a coarse supercategory prompt, such as ``mug'' or ``shoe'', but must distinguish a particular object from visually similar instances of the same category.

QART reaches $36.3\%$ PR-AUC and $61.3\%$ NDCG, improving PR-AUC by $8.4$\% over SteerViT under the same supercategory prompting setting and by $6.7$\% over unconditioned DINOv2 (Fig.~\ref{fig:rep_pods}).

The query itself cannot identify the correct instance. QART must therefore use it to determine which region or semantic content is relevant while still preserving visual cues that are more specific than the language supervision. The improvement on PODS shows that steering does not collapse the representation to only what is explicitly stated in the prompt. Instead, the query determines where representational capacity is focused while fine-grained visual information within that region is retained.

\subsection{Does Compactness Translate into Real Inference Gains?}
\label{sec:results_efficiency}

A shorter visual sequence is useful only if the reduction offsets the cost of constructing it. We therefore measure end-to-end VLM inference as batch size increases.

At batch size one, QART's recurrent tokenization introduces more overhead than is recovered by the shorter language-model sequence, resulting in a $0.87\times$ end-to-end speed ratio. The balance changes as batch size increases and processing the visual sequence inside the language model becomes more expensive. At batch size four, QART reaches $1.14\times$ end-to-end and $1.84\times$ prefill speed-up. At batch size eight, these gains increase to $1.53\times$ and $2.54\times$, respectively, with throughput increasing from $21.0$ to $32.2$ samples/s (Fig.~\ref{fig:efficiency_results}).

The different scaling of tokenization and language-model processing explains this trend. QART incurs a largely fixed cost to construct its representation, while reducing the sequence subsequently processed by the 7B language model. Its advantage therefore becomes larger when downstream sequence processing dominates the computation.

\begin{insightbox}
\textbf{QART trades recurrent tokenization overhead for substantially cheaper downstream sequence processing.}
The benefit grows under batched inference, reaching $1.53\times$ end-to-end and $2.54\times$ prefill speed-up at batch size eight.
\end{insightbox}

\subsection{What Does Query Conditioning Control?}
\label{sec:query_conditioning_ablation}

Query conditioning enters QART in two places.
It determines the visual support available during recurrent refinement, and it
defines the evidence used by the  reconstruction coverage $c^{(r)}$ to control slot growth and
halting.
We ablate these two roles on CORE, and illustrate the results in Tab.~\ref{tab:query_conditioning_ablation}.

Making the budget decisions query-agnostic has almost no effect on retrieval
accuracy, but removes the computational adaptivity of QART.
The model reaches $92.08\%$ P@1, yet activates all $32$ slots and runs all six
rounds.
Using query-relevant evidence for growth and halting preserves essentially the
same accuracy while reducing the average budget to $17.25$ slots and $4.25$
rounds.
This corresponds to roughly $46\%$ fewer active tokens and $29\%$ fewer
recurrent rounds.

The query-relevant support itself plays a different role.
When the relevance threshold $\tau$ is set to zero, all patches participate rather
than focusing recurrence on the queried evidence.
When there is no query-conditioning for the visual support, P@1 drops to $79.7\%$, still outperforming the SigLIP2 number.
Thus, query conditioning of the visual support is important for steering the
representation, whereas query conditioning of the reconstruction coverage
enables efficient allocation of capacity.

\begin{insightbox}
\textbf{Query conditioning serves two complementary roles in QART.}
It enhances steering of the representation toward the requested visual evidence, while
query-aware growth and halting avoid spending capacity on information that the
query does not require.
\end{insightbox}

\begin{table}[t]
    \centering
    \caption{
        Ablation of the two roles of query conditioning in QART on CORE.
        Query conditioning helps steer the visual support used during recurrence
        and independently control the evidence used for slot growth and halting.
    }
    \label{tab:query_conditioning_ablation}
    \small
    \begin{tabular}{ccccc}
        \toprule
        Query-steered support
        & Query-aware budget
        & P@1 $\uparrow$
        & Mean $K$ $\downarrow$
        & Mean rounds $\downarrow$ \\
        \midrule
        \cmark & \cmark & 92.01 & 17.25 & 4.25 \\
        \cmark & \xmark & 92.08 & 32.00 & 6.00 \\
        \xmark & \xmark & 79.7 & 32 & 6.00 \\
        \bottomrule
    \end{tabular}
\end{table}

%% file: sections/conclusion.tex
\section{Conclusion}
\label{sec:conclusion}

We introduced QART, a recurrent visual tokenizer that combines query steerability with budget adaptivity. QART steers its representation toward query-relevant evidence and allocates additional visual tokens only when that evidence is not yet represented with sufficient fidelity. Across retrieval, grounding, compositional matching, and visual question answering, QART produces compact representations with substantially fewer tokens while preserving strong downstream performance. Its gains are strongest when predictions depend on localized visual evidence, while on general VQA it remains competitive with existing token-reduction methods. Overall, these results show that visual representations need not be fixed in either content or capacity, and can instead be shaped by what each query requires.

%% file: sections/ai_use_statement.tex
\section*{AI Use Statement}
We have used generative AI tools for providing feedback on experiments, reformatting datasets, designing some experiments, modifying some figures, editing parts of the research paper and editing some parts of the software code. We have not used generative AI tools to  generate synthetic data sets, help develop theoretical models or conceptual frameworks, formulate mathematical claims, provide critical ingredients for proving mathematical claims, assist in the writing of proofs, propose or refine hypotheses, design research  methodology, implement methods, assist with translation, support qualitative and thematic data analysis, interpret results. We take responsibility for the final content of this work, including text, claims or artifacts produced with the aid of generative AI.

%% file: sections/ethics_statement.tex
\section*{Ethics Statement}

This work uses publicly available datasets and pretrained models and does not involve new
human-subject data collection. As with other vision and vision--language systems, QART may inherit
biases and limitations from its training data and pretrained encoders. Any downstream deployment
should therefore consider the provenance and biases of the underlying data and models, as well as
the requirements of the target application.

%% file: sections/reproducibility_statement.tex
\section*{Reproducibility Statement}

To the best of my knowledge, all the necessary details needed to reproduce the results have been provided in the main paper and appendix. However, the code cannot be released as this projected was conducted with an industrial partner which limits us. However, if absolutely necessary, we can try to release the necessary supplementary materials.

%% file: sections/appendix.tex

\section{Implementation Details}
\label{app:implementation}

\subsection{Datasets}
\label{app:datasets}

\paragraph{Tokenizer pretraining data.}
We pretrain QART on a mixture of referring-expression, scene-understanding,
and region-description datasets comprising RefCOCO \cite{yu2016refcoco}, RefCOCO+ \cite{yu2016refcoco}, RefCOCOg \cite{yu2016refcoco},
gRefCOCO \cite{liu2023gres}, Mapillary Vistas \cite{neuhold2017mapillary}, and Visual Genome (VG) \cite{krishna2017visual}.
The complete mixture contains $2{,}078{,}367$ image--text annotation records;
this number counts annotations rather than unique images.
Table~\ref{tab:mixture} summarizes the sampling distribution.
Training examples are sampled with replacement according to the specified
mixture weights.

\begin{table}[t]
    \centering
    \caption{
        Data mixture used for QART pretraining.
        VG is separated into native Visual Genome images and images overlapping
        with COCO.
    }
    \label{tab:mixture}
    \small
    \begin{tabular}{lcc}
        \toprule
        Source & Sampling probability & Validation records \\
        \midrule
        RefCOCO & 0.10 & 10,834 \\
        RefCOCO+ & 0.10 & 10,758 \\
        RefCOCOg & 0.10 & 9,536 \\
        gRefCOCO & 0.20 & 14,229 \\
        Mapillary Vistas & 0.15 & 3,053 \\
        VG, native images & 0.33 & 3,003 \\
        VG, COCO images & 0.02 & 381 \\
        \bottomrule
    \end{tabular}
\end{table}

\paragraph{Query-relevance supervision.}
During tokenizer pretraining, we use dense query-region targets to supervise the relevance
predictor, following Ruthardt et al.~\citep{ruthardt2026steervit}.
For the COCO-family referring-expression datasets~\citep{yu2016refcoco},
relevance targets are derived from segmentation annotations.
For Mapillary Vistas~\citep{neuhold2017mapillary} and Visual Genome~\citep{krishna2017visual},
we use cached dense query-region alignment maps.
The resulting heatmaps are retained as soft targets in $[0,1]$ rather than being
converted to binary bounding boxes. These targets provide training supervision
for the relevance predictor and the query-weighted reconstruction objective, but
are not used to define the visual support or the recurrent growth and halting
decisions. The recurrent tokenizer instead operates on the support predicted by
the relevance module,
$\mathcal{A}=\{i\mid p_i>\tau\}$,
both during training and inference.

\paragraph{Vision--language alignment data.}
For integration with LLaVA-1.5-7B \cite{liu2023LLaVA}, we train the visual projector in two stages.
The first stage performs one epoch of Visual Genome-based alignment using
native-image VG region descriptions.
The region phrase conditions QART, while the language model receives a generic
region-description instruction and predicts the corresponding phrase.

The second stage trains the projector for one epoch on the full
LLaVA-Instruct-665K mixture \cite{liu2023LLaVA}.
QART and the language model remain frozen throughout both stages, and only the visual projector is optimized.

\subsection{Setup}
\label{app:setup}

\paragraph{Backbone and input representations.}
Our main configuration uses the frozen vision and text encoders from
\texttt{google/siglip2-so400m-patch14-384}.
Images are processed at $384\times384$ resolution.
The visual encoder produces a $27\times27$ grid of $N=729$ patch features,
each with dimension $1152$.
We precompute and cache these frozen visual features in FP16.

A bias-free linear layer projects the visual features to the tokenizer
dimension $D=768$.

The frozen text encoder provides token-level query representations, which are
projected to the same $768$-dimensional space when necessary.
Padding tokens are masked, and a masked mean of the query features is used to
initialize the slot bank.
Tokenizer pretraining uses a maximum text length of 32 tokens.

The reconstruction target is the projected visual feature map used by the tokenizer. Specifically, the frozen backbone features are first projected to the tokenizer dimension, and the decoder reconstructs these projected visual features rather than RGB pixels.

\paragraph{Relevance predictor.}
The relevance predictor consists of five visual-to-text cross-attention layers
with eight attention heads. Visual patches act as queries and text tokens as
keys and values. Each layer uses pre-layer normalization~\citep{nguyen2019transformers}
and a residual connection without an intermediate feed-forward block.
A final linear projection followed by a sigmoid predicts one relevance score
$p_i \in [0,1]$ for each visual patch.

Queries and keys are normalized independently within each attention head after
their learned projections. At inference, the query-relevant visual support is
defined as
\begin{equation}
    \mathcal{A}
    =
    \{i \mid p_i > \tau\}.
    \label{eq:app_relevance_support}
\end{equation}
The relevance set is computed once before recurrence and remains fixed across
recurrent rounds. For tasks in which a visual target is guaranteed to exist,
an empty selection falls back to the 100 patches with the highest relevance
scores.

\paragraph{Recurrent tokenizer.}
QART maintains at most $K_{\max}=32$ active latent slots. A masked mean of the query-token representations produces a pooled query representation $\bar{\mathbf{q}}$. At initialization, a learned slot projector $f_{\mathrm{init}}$ maps $\bar{\mathbf{q}}$ to $K_0=2$ query-conditioned slots,
\begin{equation}
\mathbf{S}^{(0)}
=
f_{\mathrm{init}}(\bar{\mathbf{q}})
\in
\mathbb{R}^{K_0 \times D}.
\end{equation}
The projector produces distinct slot embeddings from the shared query representation, followed by per-slot layer normalization.

Each recurrent round uses a three-layer pre-normalized Transformer with eight attention heads and a feed-forward width of 3072. The complete three-layer Transformer is shared across recurrent rounds, while the three layers within the block use distinct parameters. Learned modality embeddings distinguish slot, visual, and text tokens, and a learned round embedding is added to the active slots.

At round $r$, the shared Transformer jointly processes the current slots
$\mathbf{S}^{(r-1)}$, the selected query-relevant visual features
$\mathbf{V}_{\mathcal{A}}$, and the text features $\mathbf{T}$:
\begin{equation}
\tilde{\mathbf{S}}^{(r)}
=
F_{\theta}
\left(
\mathbf{S}^{(r-1)},
\mathbf{V}_{\mathcal{A}},
\mathbf{T}
\right).
\end{equation}
A subsequent slot-to-patch cross-attention readout allows the resulting slots to gather additional information from the selected visual evidence:
\begin{equation}
\mathbf{S}^{(r)}
=
\mathrm{LN}
\left(
\tilde{\mathbf{S}}^{(r)}
+
\mathrm{MHA}
\left(
\tilde{\mathbf{S}}^{(r)},
\mathbf{V}_{\mathcal{A}},
\mathbf{V}_{\mathcal{A}}
\right)
\right).
\end{equation}
Only the slot states $\mathbf{S}^{(r)}$ are propagated recurrently. The selected visual features $\mathbf{V}_{\mathcal{A}}$ and text features $\mathbf{T}$ remain fixed and are supplied again at every round.

The joint self-attention uses per-head QK normalization. We additionally use shared learned residual-scaling parameters
\begin{equation}
\mathbf{U}
=
\alpha \mathbf{X}
+
\beta\,\mathrm{SA}\!\left(\mathrm{LN}(\mathbf{X})\right),
\end{equation}
\begin{equation}
\mathbf{X}'
=
\alpha \mathbf{U}
+
\beta\,\mathrm{FFN}\!\left(\mathrm{LN}(\mathbf{U})\right),
\end{equation}
initialized to $\alpha=0.75$ and $\beta=0.25$.

When the current slot set does not represent the selected visual evidence with sufficient fidelity, $\Delta K=6$ additional slots are initialized from $\bar{\mathbf{q}}$ using $f_{\mathrm{init}}$ and concatenated with the existing slots before the next recurrent round. The active slot count therefore follows
\begin{equation}
K_{r+1}
=
\begin{cases}
\min(K_r+\Delta K,K_{\max}),
&
c^{(r)} < \eta_{\mathrm{grow}},\\
K_r,
&
\text{otherwise}.
\end{cases}
\end{equation}

\paragraph{Reconstruction and adaptive computation.}
A shared decoder reconstructs the dense visual feature grid from the active slots after each recurrent round. The decoder uses $N$ learned positional queries and four pre-normalized cross-attention/FFN blocks, with the active slots supplying keys and values. Each block uses eight attention heads and a feed-forward width of $4D$. A final layer normalization and linear projection produce the reconstructed visual features
$\hat{\mathbf{V}}^{(r)} \in \mathbb{R}^{N \times D}$.

For visual feature $\mathbf{v}_i \in \mathbb{R}^{D}$ and its reconstruction
$\hat{\mathbf{v}}^{(r)}_i$ at round $r$, we compute the normalized squared reconstruction error as
\begin{equation}
\frac{1}{D}
\left\|
\hat{\mathbf{v}}_i^{(r)}-\mathbf{v}_i
\right\|_2^2.
\label{eq:app_reconstruction_error}
\end{equation}
Representation completion at round $r$ is measured over the selected query-relevant support,
\begin{equation}
    c^{(r)}
    =
    \frac{1}{|\mathcal{A}|}
    \sum_{i\in\mathcal{A}}
    \mathbb{I}
    \left[
       \frac{1}{D}
\left\|
\hat{\mathbf{v}}_i^{(r)}-\mathbf{v}_i
\right\|_2^2  < \delta
    \right].
\end{equation}

We use $\delta=0.30$,
$\eta_{\mathrm{grow}}=0.85$, and
$\eta_{\mathrm{halt}}=0.95$.
If $c^{(r)}<\eta_{\mathrm{grow}}$, up to six additional slots are activated,
subject to the $32$-slot limit.
The resulting maximum-capacity sequence is
$2,8,14,20,26,32$.
Inference may halt after a minimum of two recurrent rounds once
$c^{(r)}\geq\eta_{\mathrm{halt}}$, and otherwise continues for at most six
rounds.
During tokenizer pretraining, all six rounds are executed while slot growth
remains adaptive.

\paragraph{Tokenizer optimization.}
We optimize the trainable QART components using AdamW with learning rate
$10^{-4}$, $(\beta_1,\beta_2)=(0.9,0.999)$, and weight decay $0.01$.
The learning rate increases linearly for 4,000 optimizer steps and then follows
a cosine schedule over 25 epochs.
Training uses a global batch size of 256, BF16 mixed precision, and
global-norm gradient clipping at $0.5$.

The training objective combines query-weighted feature reconstruction and
relevance prediction as described in Sec.~\ref{sec:objective}.
We use
$\lambda_{\mathrm{rec}}=1$ and
$\lambda_{\mathrm{rel}}=0.5$.
The implementation additionally includes a slot-count term with weight $0.01$.
Because slot allocation is discrete, this term does not propagate gradients
through the allocation decisions.

\paragraph{Vision--language projector.}
For LLaVA-1.5-7B, the active QART slots are mapped to the language-model
embedding dimension using a two-layer projector
\begin{equation}
    768
    \rightarrow
    4096
    \rightarrow
    4096,
\end{equation}
with input layer normalization, GELU activation, and output layer
normalization.

Both projector-training stages use AdamW with learning rate
$2\times10^{-4}$, zero weight decay, a 500-step linear warmup followed by
cosine decay, and gradient-norm clipping at $1.0$.
The global batch size is 32.
The first Visual Genome alignment stage is run for one epoch, followed by one
epoch on LLaVA-Instruct-665K.
Only the projector parameters are updated in both stages.

\subsection{Evaluation Protocols}
\label{app:evaluation}

\paragraph{Shared representation evaluation.}
CORE, GeneCIS, PODS, and Winoground use the frozen pretrained QART tokenizer
without benchmark-specific fitting.
For retrieval-style evaluations, we mean-pool the active slots, compute the
representation in FP32, apply $\ell_2$ normalization, and use cosine
similarity.

Unless stated otherwise, transfer experiments use the same SigLIP2 QART
checkpoint and the relevance threshold
$\tau=0.25$.
GeneCIS, PODS, and Winoground execute each image--query pair independently so
that recurrent halting is determined per example.

\paragraph{CORE.}
We evaluate on the \texttt{JonaRuthardt/CORE} test split using the provided
scene-specific galleries and object prompts.
For each target object, all images in its gallery are encoded under the same
query.
Each positive image is used as a retrieval query, while itself and variants of
the same source image are excluded from the gallery.
We report precision at rank one (P@1) over the eligible retrieval queries.

\paragraph{GeneCIS.}
We evaluate the GeneCIS \cite{vaze2023genecis} \emph{Focus Object} split.
For every query, the reference image and its 15 candidate images are encoded
using the same object condition.
Candidate membership from the benchmark is preserved without introducing
additional negatives.
Candidates are ranked using cosine similarity between normalized mean-slot
representations, and we report R@1, R@2, and R@3.

\paragraph{VQAv2.}
For VQAv2 \cite{goyal2017vqav2}, we use the QART--LLaVA-1.5-7B model obtained after the two-stage
projector training described above.
QART and the 7B language model remain frozen, and the relevance threshold is set to $\tau=0.05$.
Only active QART slots are projected into the language-model embedding space
and inserted before the textual question.

Answers are generated autoregressively using greedy decoding.
We use at most 16 generated tokens and evaluate predictions using the standard
VQAv2 evaluation protocol.

For VQAv2, we reproduce LLaVA+SigLIP2 where the visual representations are replaced with siglip tokens, and Q-Former \cite{li2023blip2} by following the same protocol as QART. The other numbers are reported from their corresponding papers. 

\paragraph{Visual7W.} \label{app:vis7w_eval}
For Visual7W, we evaluate it on the pointing task. For the pointing task, the 4 candidate bounding boxes are given in [x1, y1, x2, y2] format. Also, the 4 candidates are given as A, B, C and D to the LLM, and the answer is always one of the 4 alphabets. Evaluation is normal accuracy checking for a perfect match. 

We reproduce all compared methods under a common evaluation
pipeline.
We use the LLaVA+SigLIP2 configuration as the base model for all the pruning baselines which is finetuned with the same protocol as qart, which is one epoch of vg-alignment and 6 epochs of Visual7W projector finetuning. And for all learnable baselines, we additionally finetune for 4 epochs on Visual7W data to adapt to the compressed representations.

\paragraph{Winoground.}
We evaluate all 400 official Winoground \cite{thrush2022winoground} examples and score all four
image--caption pairings.
For an image $I_i$ conditioned on caption $c_j$, QART reconstructs the visual
feature grid from its active slots.
We then apply the frozen SigLIP2 attention pooler and compute cosine similarity
with the native SigLIP2 text representation,
\begin{equation}
    S_{ij}
    =
    \cos
    \left(
        H_v\bigl(\hat V(I_i,c_j)\bigr),
        H_t(c_j)
    \right).
\end{equation}
We compute the official Text, Image, and Group accuracies from the resulting
$2\times2$ similarity matrix.
For the fixed-query control, the same generic text query is used to condition
QART for every image while the official Winoground scoring protocol remains
unchanged.

\paragraph{PODS.}
We follow the PODS \cite{sundaram2025personalized} evaluation protocol over 100 target objects, 300 reference
images, and 10,888 test images.
For the main experiment, the dataset's coarse supercategory label is used as
the conditioning query for all reference and test images associated with a
target object.

For classification, each test image is scored by its maximum cosine similarity
to the three positive reference representations.
We compute average precision independently for each target object and report
the macro average.
For retrieval, each positive test image ranks all 300 references, with the
three references belonging to the same target object treated as relevant.
We compute binary NDCG over the complete ranking and average first within each
target object and then across objects.

\section{Evaluating Backbone generalization with DINOv2}
\label{app:additional_results}

\subsection{Data and Setup}
\label{app:dinov2_setup}

We additionally train QART using frozen DINOv2-B/14 visual features and a
frozen RoBERTa-large text encoder.
Images are resized to $378\times378$, producing a $27\times27$ grid of
$729$ patch features with dimension $768$ after removing the CLS token.
RoBERTa-large provides the query features, which are projected to the same
$768$-dimensional tokenizer space.

The remaining QART architecture is unchanged from the SigLIP2 configuration.
We use the same relevance predictor, recurrent tokenizer, reconstruction
decoder, slot-allocation policy, training objective, data mixture, optimizer,
and training schedule described in Sec.~\ref{app:implementation}.
The model uses two initial slots, a maximum of $32$ slots, and at most six
recurrent rounds.

The DINOv2 configuration differs from the main SigLIP2 model in an important
respect.
SigLIP2 jointly pretrains its visual and textual representations, whereas
DINOv2 and RoBERTa are independently pretrained and do not share a
cross-modal embedding space.
Consequently, the DINOv2--RoBERTa configuration provides a weaker
initialization for query-conditioned visual representation learning despite
using the same QART architecture.

\paragraph{VQAv2 setup.}
For VQAv2 \cite{goyal2017vqav2}, we follow the same protocol as followed for the SigLIP2 QART variant.

\subsection{Results}
\label{app:dinov2_eval}

\paragraph{CORE.}
We evaluate DINOv2 QART using the same CORE protocol as the SigLIP2 model.
The relevance threshold is applied to the predicted patch-relevance scores,
and image representations are obtained by mean pooling the active slots,
followed by $\ell_2$ normalization and cosine similarity.

At $\tau=0.25$, DINOv2 QART reaches $90.33\%$ P@1 with
$23.34$ active slots on average.
Its best recorded operating point reaches $90.59\%$ P@1 at
$\tau=0.20$.
For comparison, the SigLIP2 QART configuration reaches $92.01\%$ P@1 at
$\tau=0.25$.

The relatively small gap on CORE indicates that the adaptive tokenization
mechanism transfers effectively to DINOv2 features, despite the absence of a
jointly pretrained vision--language representation.

\paragraph{VQAv2.}
We evaluate the DINOv2 variant using the same LLaVA-1.5-7B integration and
projector-training protocol as the SigLIP2 QART model. The results are shown in Tab.~\ref{tab:dinov2_vqav2}.

\begin{table}[t]
    \centering
    \caption{
        VQAv2 results for QART with SigLIP2 and DINOv2 visual backbones.
        Both variants use the same LLaVA-1.5-7B integration and projector-training
        protocol.
    }
    \label{tab:dinov2_vqav2}
    \small
    \begin{tabular}{lcc}
        \toprule
        Metric & SigLIP2 QART & DINOv2 QART \\
        \midrule
        Overall accuracy & \textbf{76.37\%} & 72.93\% \\
        Yes/No & \textbf{91.20\%} & 89.87\% \\
        Number & \textbf{60.80\%} & 54.16\% \\
        Other & \textbf{69.21\%} & 65.01\% \\
        \midrule
        Mean active slots & \textbf{29.45} & 30.96 \\
        \bottomrule
    \end{tabular}
\end{table}

Under the matched training protocol, the DINOv2 variant reaches
$72.93\%$ overall accuracy compared with $76.37\%$ for SigLIP2 QART.
The largest difference appears on number questions, while the gap is smaller
for yes/no questions.
DINOv2 also uses slightly more active slots on average.

Although the DINOv2 variant trails SigLIP2, its VQAv2 performance remains
competitive with prior visual-token pruning and reduction methods operating at
similarly aggressive token budgets, as reported in the main paper.
This indicates that the QART formulation remains effective even when the visual
and textual encoders are not jointly pretrained.

We attribute the remaining gap primarily to the weaker cross-modal initialization
of the independently pretrained DINOv2 and RoBERTa encoders, which lack the
native vision--language alignment available in SigLIP2.

\subsection{Discussion}
\label{app:dinov2_discussion}

The DINOv2 results show that QART is not tied to a particular visual backbone.
The variant remains close to SigLIP2 on CORE and retains competitive VQAv2
performance at a compact visual-token budget, despite using independently
pretrained vision and text encoders.

The stronger SigLIP2 results are consistent with the benefit of native
cross-modal alignment.
SigLIP2 jointly learns visual and textual representations, whereas DINOv2 and
RoBERTa are pretrained independently and therefore provide a weaker starting
point for query-conditioned representation learning.
We therefore view the DINOv2 experiments as evidence that QART generalizes
across backbone families, while also highlighting the importance of aligned
vision--language features for its strongest performance.

\section{Additional Results and Analyses for QART}

\subsection{Threshold Sensitivity and Adaptive Budgeting}
\label{app:threshold_sensitivity}

The relevance threshold $\tau$ controls the spatial support
selected for recurrent tokenization.
Changing this threshold therefore affects both \emph{which} image regions are
represented and the amount of representational capacity subsequently allocated
by QART.
We analyze this interaction on CORE, GeneCIS, and Winoground using the same
SigLIP2 QART model.

Table~\ref{tab:threshold_sensitivity} reports the primary metric for each task
together with the resulting mean active-slot count when available.
For CORE, we report P@1; for GeneCIS, R@1; and for Winoground, Group accuracy.

\begin{table}[t]
    \centering
    \caption{
        Sensitivity to the relevance threshold.
        Lower thresholds generally increase the selected visual support, but
        performance does not vary monotonically with the resulting token budget.
        Mean $K$ denotes the average number of active QART slots.
    }
    \label{tab:threshold_sensitivity}
    \small
    \begin{tabular}{lccc}
        \toprule
        Task & $\tau$ & Primary metric $\uparrow$ & Mean $K$ \\
        \midrule

        \multirow{5}{*}{CORE}
        & 0.10 & 85.61 & -- \\
        & 0.20 & 90.96 & -- \\
        & 0.25 & \textbf{92.01} & 17.25 \\
        & 0.35 & 91.92 & -- \\
        & 0.50 & 89.79 & -- \\

        \midrule

        \multirow{6}{*}{GeneCIS}
        & 0.05 & 12.81 & 30.04 \\
        & 0.10 & 13.32 & 28.04 \\
        & 0.20 & \textbf{15.10} & 25.38 \\
        & 0.25 & 13.32 & 24.43 \\
        & 0.35 & 14.18 & 22.63 \\
        & 0.50 & 12.91 & 21.52 \\

        \midrule

        \multirow{5}{*}{Winoground}
        & 0.10 & 15.25 & 31.07 \\
        & 0.20 & 15.25 & 28.84 \\
        & 0.25 & \textbf{19.25} & 27.49 \\
        & 0.35 & 17.00 & 25.40 \\
        & 0.50 & 15.00 & 22.18 \\

        \bottomrule
    \end{tabular}
\end{table}

The optimal operating point differs across tasks.
CORE performs best at $\tau=0.25$, whereas GeneCIS reaches its
highest R@1 at $\tau=0.20$.
Winoground also favors an intermediate threshold rather than either extreme.

More importantly, increasing the selected visual support and the resulting
token budget does not guarantee improved performance.
On GeneCIS, lowering the threshold from $0.20$ to $0.05$ increases the mean
slot count from $25.38$ to $30.04$, yet R@1 decreases from $15.10\%$ to
$12.81\%$.
A similar pattern appears on Winoground, where
$\tau=0.10$ uses $31.07$ slots on average but obtains lower
Group accuracy than $\tau=0.25$, which uses only $27.49$ slots.

These results highlight that QART's budget should not be interpreted as a
simple ``more tokens is better'' trade-off.
The relevance threshold jointly controls the spatial extent and composition of
the selected evidence, while the recurrent tokenizer allocates capacity to
represent that evidence.
Different tasks therefore favor different combinations of visual coverage and
token budget.

\begin{insightbox}
    \textbf{Effective visual representations require adapting both spatial support
and representational capacity.}
\end{insightbox}
Broader coverage generally increases the available token budget, but additional
tokens do not monotonically improve downstream performance.
The task-dependent optima suggest that a single fixed visual budget or uniformly
holistic representation is unlikely to be optimal across different visual
reasoning settings.

\subsection{Does Recurrence Refine the Representation?}
\label{sec:recurrence_analysis}

\begin{figure*}[t]
    \centering
    \includegraphics[width=\textwidth]{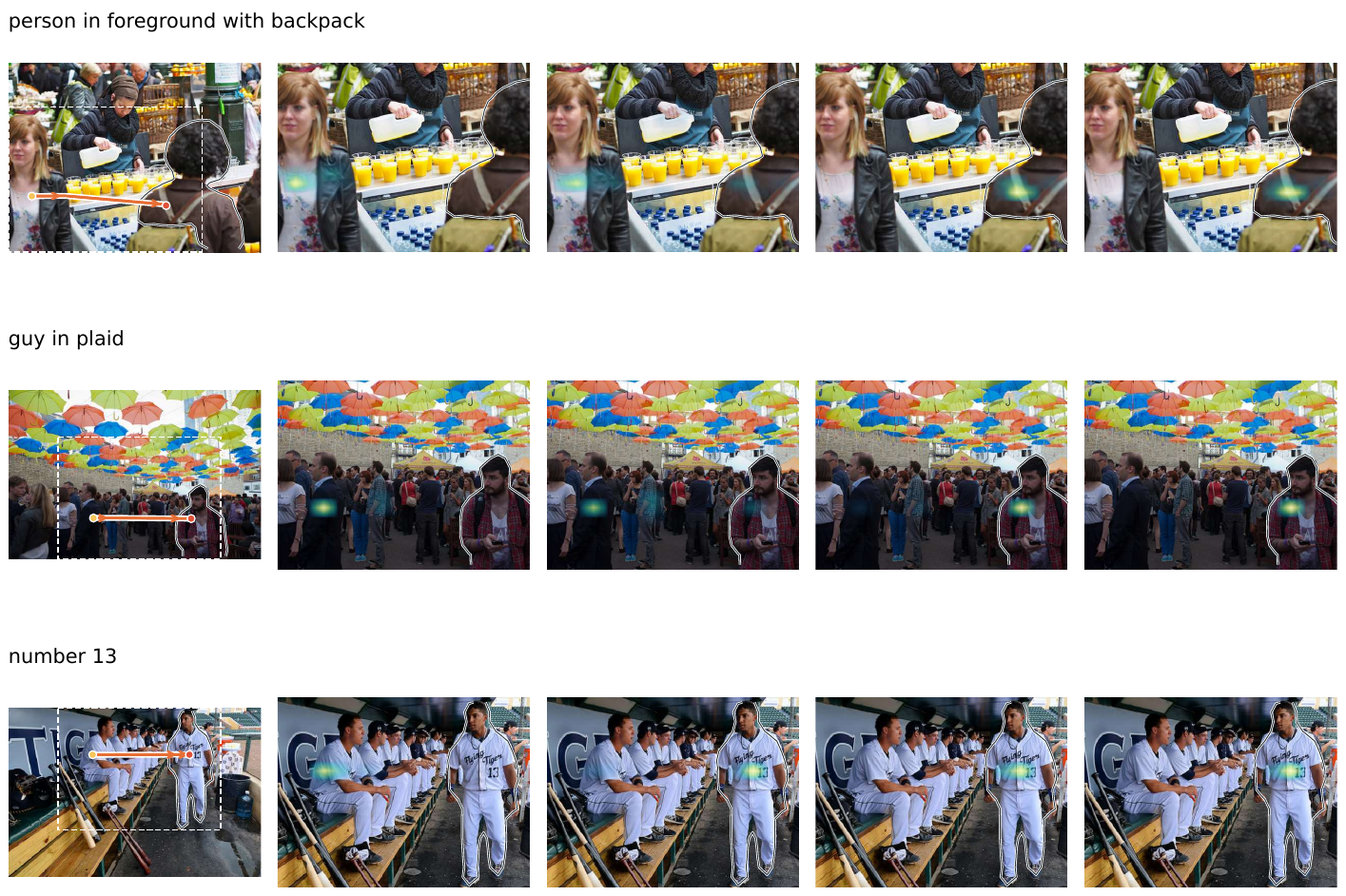}
    \caption{
        \textbf{Representation refinement across recurrent rounds.}
        Each row shows the spatial support of an active QART slot over successive
        rounds for a different query.
        The support progressively shifts toward the query-relevant region, including
        the foreground backpack, the person in plaid, and jersey number 13.
    }
    \label{fig:recurrence_refinement}
\end{figure*}

QART does not construct its representation in a single step.
The active slots are repeatedly refined using the same query and visual
evidence, allowing their spatial support to change across recurrent rounds.
Figure~\ref{fig:recurrence_refinement} shows representative examples of this
behavior.

The progression is particularly clear when the initial representation attends
to a nearby distractor or only coarsely identifies the relevant region.
For the query ``person in foreground with backpack,'' the early support is
spread over nearby foreground content before progressively concentrating on
the queried person and backpack.
A similar shift occurs for ``guy in plaid,'' where the support moves toward
the queried individual over successive rounds.
For ``number 13,'' the representation initially responds to nearby players
before becoming localized around the jersey number of the target player.

These examples suggest that recurrence provides more than additional
representational capacity.
Later rounds can revise where the active slots attend, progressively moving
their support toward evidence that is better aligned with the query.
This is useful when the relevant region is visually ambiguous or surrounded
by similar content, where a single tokenization step may commit too early to
an incorrect or overly broad region.

\begin{insightbox}
\textbf{Recurrence allows QART to revise, rather than simply expand, its
representation.}
Across rounds, slot support can move from coarse or distracting regions toward
the queried visual evidence, providing a mechanism for iterative
query-conditioned refinement.
\end{insightbox}

\subsection{Is QART query-steerable and budget-adaptive?}
\label{sec:qualitative_adaptivity}

\begin{figure*}[t]
    \centering
    \includegraphics[width=0.82\textwidth]{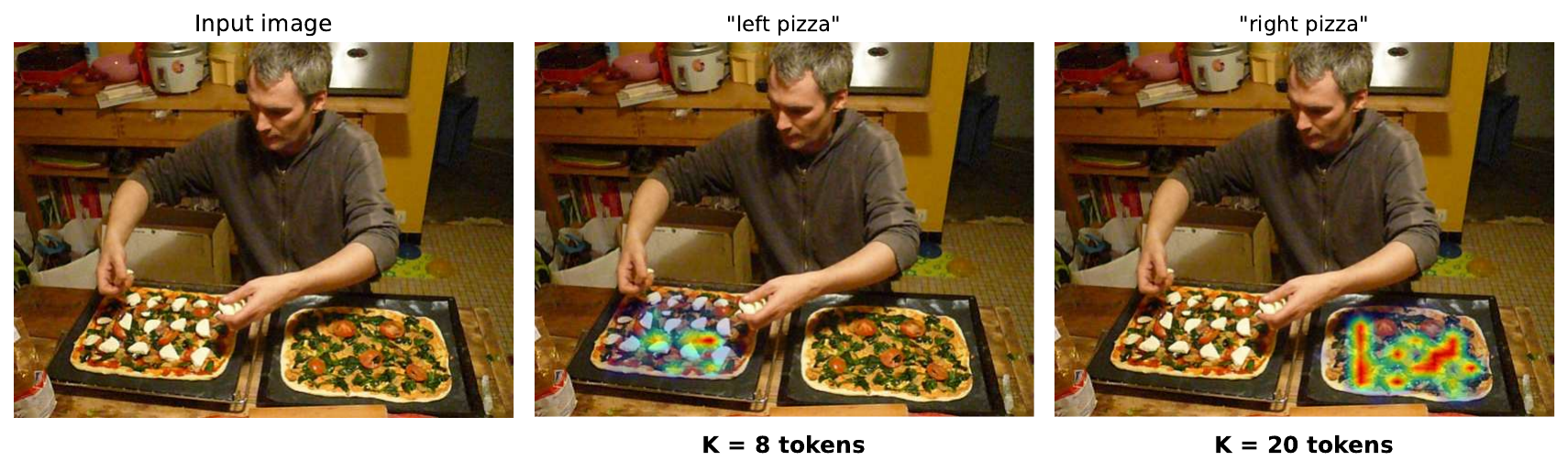}
    \caption{
        \textbf{Query-dependent visual support and token allocation.}
        For the same input image, changing the query from ``left pizza'' to
        ``right pizza'' redirects the visual support to the requested region
        and changes the active token budget from $K=8$ to $K=20$.
    }
    \label{fig:query_steerability}
\end{figure*}

Figure~\ref{fig:query_steerability} qualitatively illustrates how the
conditioning query changes the visual representation.
For the same input image, QART shifts its support from the left pizza to the
right pizza when the query changes.
Importantly, the representation does not only move spatially: the active token
budget also changes from $K=8$ to $K=20$.
This provides a direct example of query steerability and budget adaptivity
within the same image.

\begin{figure*}[t]
    \centering
    \includegraphics[width=0.96\textwidth]
    {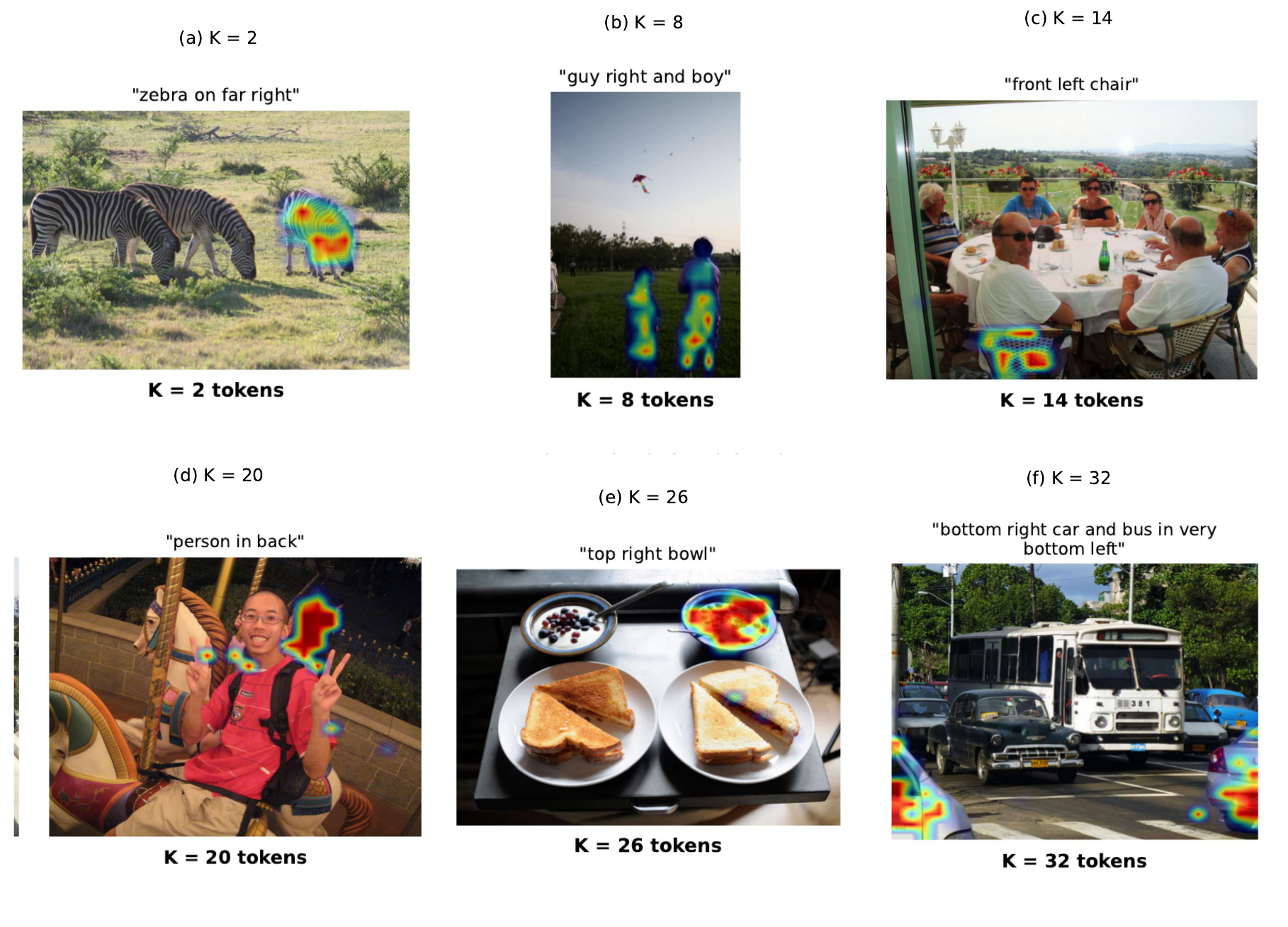}
    \caption{
        \textbf{QART adapts its token budget across image--query pairs.}
        Representative examples span the allocation range from $K=2$ to
        $K=32$. Localized queries can be represented with only a few active
        tokens, while broader or more spatially distributed visual evidence
        receives a larger representation.
    }
    \label{fig:budget_adaptivity}
\end{figure*}

The allocated capacity also varies substantially across different image--query
pairs.
Figure~\ref{fig:budget_adaptivity} shows representative examples spanning
$K=2$ to the full $K=32$ slot budget.
Localized targets such as ``zebra on far right'' and ``top right bowl'' require only
a small representation, whereas other queries receive substantially more
capacity.
Queries involving broader or spatially distributed evidence can activate most
or all of the available slots.

The token count is not determined by the number of queried objects alone.
Instead, it reflects the visual evidence that QART needs to preserve for a
particular image--query pair, including its spatial extent, appearance, and
surrounding context.
Together, Figs.~\ref{fig:query_steerability} and
\ref{fig:budget_adaptivity} show that QART adapts both the content and the
capacity of its representation rather than applying a fixed visual budget to
every query.

\begin{insightbox}
\textbf{QART adapts both where it represents and how much it represents.}
Changing the query can redirect the visual support even for the same image,
while the active token budget varies from a few slots to the full capacity
depending on the evidence required by the query.
\end{insightbox}

\subsection{Limitations}
While QART is evaluated across retrieval, compositional matching, visual question answering,
and grounding, our experiments focus on single-image settings. Extending query-adaptive
token allocation to video, multi-turn interaction, and temporally evolving visual observations
remains an interesting direction for future work.

The current formulation also uses a small number of thresholds to control relevance selection,
slot growth, and halting. As shown in our sensitivity analysis, the preferred operating point can
vary across tasks. Exploring fully learned or calibration-free allocation criteria could further
simplify deployment across heterogeneous downstream settings.

Finally, QART uses dense query-region supervision during tokenizer pretraining to train the
relevance predictor and reconstruction objective. This supervision is not required at inference,
but future work could investigate weaker or self-supervised relevance signals to reduce the
annotation requirements of tokenizer pretraining.

%% file: iclr2027_conference.bib
@inproceedings{ryoo2021tokenlearner,
  title     = {{TokenLearner}: Adaptive Space-Time Tokenization for Videos},
  author    = {Ryoo, Michael S. and Piergiovanni, A. J. and Arnab, Anurag and Dehghani, Mostafa and Angelova, Anelia},
  booktitle = {Advances in Neural Information Processing Systems},
  volume    = {34},
  year      = {2021},
  url       = {https://proceedings.neurips.cc/paper/2021/hash/6a30e32e56fce5cf381895dfe6ca7b6f-Abstract.html}
}

@article{krishna2017visual,
  title={Visual genome: Connecting language and vision using crowdsourced dense image annotations},
  author={Krishna, Ranjay and Zhu, Yuke and Groth, Oliver and Johnson, Justin and Hata, Kenji and Kravitz, Joshua and Chen, Stephanie and Kalantidis, Yannis and Li, Li-Jia and Shamma, David A and others},
  journal={International Journal of Computer Vision},
  volume={123},
  number={1},
  pages={32--73},
  year={2017},
  publisher={Springer}
}

@inproceedings{neuhold2017mapillary,
  title={The mapillary vistas dataset for semantic understanding of street scenes},
  author={Neuhold, Gerhard and Ollmann, Tobias and Rota Bulo, Samuel and Kontschieder, Peter},
  booktitle={Proceedings of the IEEE international conference on computer vision},
  pages={4990--4999},
  year={2017}
}

@inproceedings{nguyen2019transformers,
  title={Transformers without tears: Improving the normalization of self-attention},
  author={Nguyen, Toan Q and Salazar, Julian},
  booktitle={Proceedings of the 16th international conference on spoken language translation},
  year={2019}
}

@inproceedings{vaze2023genecis,
  title={Genecis: A benchmark for general conditional image similarity},
  author={Vaze, Sagar and Carion, Nicolas and Misra, Ishan},
  booktitle={2023 IEEE/CVF Conference on Computer Vision and Pattern Recognition (CVPR)},
  pages={6862--6872},
  year={2023},
  organization={IEEE}
}

@inproceedings{sundaram2025personalized,
  title={Personalized representation from personalized generation},
  author={Sundaram, Shobhita and Chae, Julia and Tian, Yonglong and Beery, Sara and Isola, Phillip},
  booktitle={International Conference on Learning Representations},
  volume={2025},
  pages={92572--92606},
  year={2025}
}

@inproceedings{liu2023gres,
  title={Gres: Generalized referring expression segmentation},
  author={Liu, Chang and Ding, Henghui and Jiang, Xudong},
  booktitle={2023 IEEE/CVF Conference on Computer Vision and Pattern Recognition (CVPR)},
  pages={23592--23601},
  year={2023},
  organization={IEEE}
}

@InProceedings{zhu2016cvpr,
  title = {{Visual7W: Grounded Question Answering in Images}},
  author = {Yuke Zhu and Oliver Groth and Michael Bernstein and Li Fei-Fei},
  booktitle = {{IEEE Conference on Computer Vision and Pattern Recognition}},
  year = 2016,
}

@inproceedings{wen2025stop,
  title={Stop looking for “important tokens” in multimodal language models: Duplication matters more},
  author={Wen, Zichen and Gao, Yifeng and Wang, Shaobo and Zhang, Junyuan and Zhang, Qintong and Li, Weijia and He, Conghui and Zhang, Linfeng},
  booktitle={Proceedings of the 2025 Conference on Empirical Methods in Natural Language Processing},
  pages={9972--9991},
  year={2025}
}

@inproceedings{li2023blip2,
  title     = {{BLIP-2}: Bootstrapping Language-Image Pre-training with Frozen Image Encoders and Large Language Models},
  author    = {Li, Junnan and Li, Dongxu and Savarese, Silvio and Hoi, Steven C. H.},
  booktitle = {Proceedings of the 40th International Conference on Machine Learning},
  series    = {Proceedings of Machine Learning Research},
  volume    = {202},
  pages     = {19730--19742},
  publisher = {PMLR},
  year      = {2023},
  url       = {https://proceedings.mlr.press/v202/li23q.html}
}

@inproceedings{radford2021learning,
  title={Learning transferable visual models from natural language supervision},
  author={Radford, Alec and Kim, Jong Wook and Hallacy, Chris and Ramesh, Aditya and Goh, Gabriel and Agarwal, Sandhini and Sastry, Girish and Askell, Amanda and Mishkin, Pamela and Clark, Jack and others},
  booktitle={International conference on machine learning},
  pages={8748--8763},
  year={2021},
  organization={PmLR}
}

@article{oquab2023dinov2,
  title={Dinov2: Learning robust visual features without supervision},
  author={Oquab, Maxime and Darcet, Timoth{\'e}e and Moutakanni, Th{\'e}o and Vo, Huy and Szafraniec, Marc and Khalidov, Vasil and Fernandez, Pierre and Haziza, Daniel and Massa, Francisco and El-Nouby, Alaaeldin and others},
  journal={arXiv preprint arXiv:2304.07193},
  year={2023}
}

@inproceedings{he2022masked,
  title={Masked autoencoders are scalable vision learners},
  author={He, Kaiming and Chen, Xinlei and Xie, Saining and Li, Yanghao and Doll{\'a}r, Piotr and Girshick, Ross},
  booktitle={2022 IEEE/CVF conference on computer vision and pattern recognition (CVPR)},
  pages={15979--15988},
  year={2022},
  organization={IEEE}
}

@inproceedings{thrush2022winoground,
  title={Winoground: Probing vision and language models for visio-linguistic compositionality},
  author={Thrush, Tristan and Jiang, Ryan and Bartolo, Max and Singh, Amanpreet and Williams, Adina and Kiela, Douwe and Ross, Candace},
  booktitle={2022 IEEE/CVF Conference on Computer Vision and Pattern Recognition (CVPR)},
  pages={5228--5238},
  year={2022},
  organization={IEEE}
}

@inproceedings{singh2022flava,
  title={Flava: A foundational language and vision alignment model},
  author={Singh, Amanpreet and Hu, Ronghang and Goswami, Vedanuj and Couairon, Guillaume and Galuba, Wojciech and Rohrbach, Marcus and Kiela, Douwe},
  booktitle={2022 IEEE/CVF Conference on Computer Vision and Pattern Recognition (CVPR)},
  pages={15617--15629},
  year={2022},
  organization={IEEE}
}

@inproceedings{hu2022scaling,
  title={Scaling up vision-language pretraining for image captioning},
  author={Hu, Xiaowei and Gan, Zhe and Wang, Jianfeng and Yang, Zhengyuan and Liu, Zicheng and Lu, Yumao and Wang, Lijuan},
  booktitle={2022 IEEE/CVF Conference on Computer Vision and Pattern Recognition (CVPR)},
  pages={17959--17968},
  year={2022},
  organization={IEEE}
}

@inproceedings{anderson2018bottomup,
  title     = {Bottom-Up and Top-Down Attention for Image Captioning and Visual Question Answering},
  author    = {Anderson, Peter and He, Xiaodong and Buehler, Chris and Teney, Damien and Johnson, Mark and Gould, Stephen and Zhang, Lei},
  booktitle = {Proceedings of the IEEE Conference on Computer Vision and Pattern Recognition},
  year      = {2018},
  pages     = {6077--6086},
  eprint    = {1707.07998},
  archivePrefix = {arXiv},
  url       = {https://www.alphaxiv.org/abs/1707.07998}
}

@inproceedings{hudson2019gqa,
  title     = {GQA: A New Dataset for Real-World Visual Reasoning and Compositional Question Answering},
  author    = {Hudson, Drew A. and Manning, Christopher D.},
  booktitle = {Proceedings of the IEEE/CVF Conference on Computer Vision and Pattern Recognition},
  year      = {2019},
  pages     = {6700--6709},
  eprint    = {1902.09506},
  archivePrefix = {arXiv},
  url       = {https://www.alphaxiv.org/abs/1902.09506}
}

@article{pantazopoulos2025visualgrounding,
  title         = {Towards Understanding Visual Grounding in Vision-Language Models},
  author        = {Pantazopoulos, Georgios and {\"O}zyi{\u{g}}it, Eda B.},
  journal       = {arXiv preprint arXiv:2509.10345},
  year          = {2025},
  eprint        = {2509.10345},
  archivePrefix = {arXiv},
  url           = {https://www.alphaxiv.org/abs/2509.10345}
}

@article{acharya2018tallyqa,
  title         = {TallyQA: Answering Complex Counting Questions},
  author        = {Acharya, Manoj and Kafle, Kushal and Kanan, Christopher},
  journal       = {arXiv preprint arXiv:1810.12440},
  year          = {2018},
  eprint        = {1810.12440},
  archivePrefix = {arXiv},
  url           = {https://www.alphaxiv.org/abs/1810.12440}
}

@inproceedings{rajabi2024gsrbench,
  title     = {GSR-BENCH: A Benchmark for Grounded Spatial Reasoning Evaluation via Multimodal LLMs},
  author    = {Rajabi, Navid and Ko{\v{s}}eck{\'a}, Jana},
  booktitle = {NeurIPS 2024 Workshop on Compositional Learning: Perspectives, Methods, and Paths Forward},
  year      = {2024},
  eprint    = {2406.13246},
  archivePrefix = {arXiv},
  url       = {https://www.alphaxiv.org/abs/2406.13246}
}

@article{pinpoint,
      title={PinPoint: Prompting with Informative Interior Points}, 
      author={Pouya Sadeghi and Shawn He and Pedro Pablo Guerrero Vela and C. Thomas and Alex Wong and Sirisha Rambhatla},
      year={2026},
      eprint={2605.26689},
      archivePrefix={arXiv},
      primaryClass={cs.CV},
      url={https://arxiv.org/abs/2605.26689}, 
}

@article{ishihara2025strideqa,
  title         = {STRIDE-QA: Visual Question Answering Dataset for Spatiotemporal Reasoning in Urban Driving Scenes},
  author        = {Ishihara, Keishi and Sasaki, Kento and Takahashi, Tsubasa and Shiono, Daiki and Yamaguchi, Yu},
  journal       = {arXiv preprint arXiv:2508.10427},
  year          = {2025},
  eprint        = {2508.10427},
  archivePrefix = {arXiv},
  url           = {https://www.alphaxiv.org/abs/2508.10427}
}

@misc{huang2026evidencerlevidenceintensivevisualreasoning,
      title={Evidence-RL: Towards Evidence-intensive Visual Reasoning},
      author={Haojie Huang and Xinlei Yu and Chengming Xu and Zhangquan Chen and Cheng Yang and Qingdong He and Yu Yang and Jiangning Zhang and Xiaobin Hu},
      year={2026},
      eprint={2608.08021},
      archivePrefix={arXiv},
      primaryClass={cs.CV},
      url={https://arxiv.org/pdf/2608.08021v1},
}

@article{yin2025dart,
  title={DART: Differentiable Adaptive Region Tokenizer for Vision Foundation Models},
  author={Shicheng Yin and Kaixuan Yin and Yang Liu and Weixing Chen and Liang Lin},
  journal={arXiv preprint arXiv:2506.10390},
  year={2025}
}

@inproceedings{yin2022avit,
        title={{A}-{V}i{T}: {A}daptive Tokens for Efficient Vision Transformer},
        author={Yin, Hongxu and Vahdat, Arash and Alvarez, Jose and Mallya, Arun and Kautz, Jan and Molchanov, Pavlo},
        booktitle={Proceedings of the IEEE/CVF Conference on Computer Vision and Pattern Recognition},
        year={2022}
    }

@article{li2026occamtoken,
  title={OccamToken: Efficient VLM Inference with Training-Free and Budget-Adaptive Token Pruning},
  author={Li, Geng and Chen, Guohao and Chen, Ting and Shan, Shilin and Zuo, Kuangji and Lyu, Bofan and An, Tuo and Li, Gen and Yang, Jianfei},
  journal={arXiv preprint arXiv:2605.29657},
  year={2026}
}

@inproceedings{kim2026and,
  title={Why and when visual token pruning fails? a study on relevant visual information shift in mllms decoding},
  author={Kim, Jiwan and Kim, Kibum and Kim, Wonjoong and Lee, Byung-Kwan and Park, Chanyoung},
  booktitle={European Conference on Computer Vision},
  pages={244--263},
  year={2026},
  organization={Springer}
}

@article{yu2024image,
  title={An image is worth 32 tokens for reconstruction and generation},
  author={Yu, Qihang and Weber, Mark and Deng, Xueqing and Shen, Xiaohui and Cremers, Daniel and Chen, Liang-Chieh},
  journal={Advances in Neural Information Processing Systems},
  volume={37},
  pages={128940--128966},
  year={2024}
}

@inproceedings{saxena2025grapheqa,
  title={GraphEQA: Using 3D Semantic Scene Graphs for Real-time Embodied Question Answering},
  author={Saxena, Saumya and Buchanan, Blake and Paxton, Chris and Chen, Bingqing and Vaskevicius, Narunas and Palmieri, Luigi and Francis, Jonathan and Kroemer, Oliver},
  booktitle={Proceedings of the Conference on Robot Learning},
  year={2025}
}

@article{cheng2024efficienteqa,
  title={EfficientEQA: An Efficient Approach for Open Vocabulary Embodied Question Answering},
  author={Cheng, Kai and Li, Zhengyuan and Sun, Xingpeng and Min, Byung-Cheol and Bedi, Amrit Singh and Bera, Aniket},
  journal={arXiv preprint arXiv:2410.20263},
  year={2024}
}

@inproceedings{liu2024grounding,
  title={Grounding dino: Marrying dino with grounded pre-training for open-set object detection},
  author={Liu, Shilong and Zeng, Zhaoyang and Ren, Tianhe and Li, Feng and Zhang, Hao and Yang, Jie and Jiang, Qing and Li, Chunyuan and Yang, Jianwei and Su, Hang and others},
  booktitle={European conference on computer vision},
  pages={38--55},
  year={2024},
  organization={Springer}
}

@inproceedings{liu2023llava,
    author      = {Liu, Haotian and Li, Chunyuan and Wu, Qingyang and Lee, Yong Jae},
    title       = {Visual Instruction Tuning},
    booktitle   = {NeurIPS},
    year        = {2023}
  }

@inproceedings{dai2023instructblip,
  title     = {{InstructBLIP}: Towards General-Purpose Vision-Language Models with Instruction Tuning},
  author    = {Dai, Wenliang and Li, Junnan and Li, Dongxu and Tiong, Anthony Meng Huat and Zhao, Junqi and Wang, Weisheng and Li, Boyang and Fung, Pascale and Hoi, Steven C. H.},
  booktitle = {Advances in Neural Information Processing Systems},
  volume    = {36},
  year      = {2023},
  doi       = {10.52202/075280-2142},
  url       = {https://proceedings.neurips.cc/paper_files/paper/2023/hash/9a6a435e75419a836fe47ab6793623e6-Abstract-Conference.html}
}

@inproceedings{hu2024mqt,
  title     = {Matryoshka Query Transformer for Large Vision-Language Models},
  author    = {Hu, Wenbo and Dou, Zi-Yi and Li, Liunian Harold and Kamath, Amita and Peng, Nanyun and Chang, Kai-Wei},
  booktitle = {Advances in Neural Information Processing Systems},
  volume    = {37},
  pages     = {50168--50188},
  year      = {2024},
  doi       = {10.52202/079017-1588},
  url       = {https://proceedings.neurips.cc/paper_files/paper/2024/hash/59c147c7d4fdb732daea3064eab949bf-Abstract-Conference.html}
}

@inproceedings{duggal2024alit,
  title     = {Adaptive Length Image Tokenization via Recurrent Allocation},
  author    = {Duggal, Shivam and Isola, Phillip and Torralba, Antonio and Freeman, William T.},
  booktitle = {International Conference on Learning Representations},
  year      = {2025},
  url       = {https://proceedings.iclr.cc/paper_files/paper/2025/hash/534ec8dfc0312d96e1c1b357304fadc2-Abstract-Conference.html}
}

@inproceedings{xiao2024flair,
  title     = {{FLAIR}: {VLM} with Fine-Grained Language-Informed Image Representations},
  author    = {Xiao, Rui and Kim, Sanghwan and Georgescu, Mariana-Iuliana and Akata, Zeynep and Alaniz, Stephan},
  booktitle = {Proceedings of the IEEE/CVF Conference on Computer Vision and Pattern Recognition},
  pages     = {24884--24894},
  year      = {2025},
  doi       = {10.1109/CVPR52734.2025.02317},
  url       = {https://openaccess.thecvf.com/content/CVPR2025/html/Xiao_FLAIR_VLM_with_Fine-grained_Language-informed_Image_Representations_CVPR_2025_paper.html}
}

@article{thirukovalluru2025tie,
  title         = {Text-Guided Semantic Image Encoder},
  author        = {Thirukovalluru, Raghuveer and Han, Xiaochuang and Dhingra, Bhuwan and Dinan, Emily and Elbayad, Maha},
  journal       = {arXiv preprint arXiv:2511.20770},
  year          = {2025},
  eprint        = {2511.20770},
  archivePrefix = {arXiv},
  url           = {https://arxiv.org/abs/2511.20770}
}

@article{ruthardt2026steervit,
  title         = {Steerable Visual Representations},
  author        = {Ruthardt, Jona and Gaur, Manu and Ramanan, Deva and Tapaswi, Makarand and Asano, Yuki M.},
  journal       = {arXiv preprint arXiv:2604.02327},
  year          = {2026},
  eprint        = {2604.02327},
  archivePrefix = {arXiv},
  url           = {https://arxiv.org/abs/2604.02327}
}

@article{li2026qmop,
  title         = {{QMoP}: Query Guided Mixture-of-Projector for Efficient Visual Token Compression},
  author        = {Li, Zhongyang and Li, Yaqian and Fang, Faming and Takezoe, Rinyoichi and Bo, Zi-Hao and Qian, Cheng and Guang, Mo and Zhang, Guixu and Long, Kaiwen},
  journal       = {arXiv preprint arXiv:2603.21232},
  year          = {2026},
  eprint        = {2603.21232},
  archivePrefix = {arXiv},
  url           = {https://arxiv.org/abs/2603.21232}
}

@inproceedings{bolya2023tome,
  title     = {Token Merging: Your {ViT} but Faster},
  author    = {Bolya, Daniel and Fu, Cheng-Yang and Dai, Xiaoliang and Zhang, Peizhao and Feichtenhofer, Christoph and Hoffman, Judy},
  booktitle = {International Conference on Learning Representations},
  year      = {2023},
  url       = {https://arxiv.org/abs/2210.09461}
}

@article{li2025qgvtc,
  title         = {{QG-VTC}: Question-Guided Visual Token Compression in {MLLMs} for Efficient {VQA}},
  author        = {Li, Shuai and Xu, Jian and Li, Xiao-Hui and Deng, Chao and Huang, Lin-Lin},
  journal       = {arXiv preprint arXiv:2504.00654},
  year          = {2025},
  eprint        = {2504.00654},
  archivePrefix = {arXiv},
  url           = {https://arxiv.org/abs/2504.00654}
}

@inproceedings{gao2026quietprune,
  title     = {{QuietPrune}: Query-Guided Early Token Pruning for Vision-Language Models},
  author    = {Gao, Tianxiao and Zhao, Shanwei and Fang, Shuo and Zhu, Shiai and Ma, Chenguang},
  booktitle = {Proceedings of the IEEE/CVF Conference on Computer Vision and Pattern Recognition},
  pages     = {3553--3562},
  year      = {2026},
  url       = {https://openaccess.thecvf.com/content/CVPR2026/html/Gao_QuietPrune_Query-Guided_Early_Token_Pruning_for_Vision-Language_Models_CVPR_2026_paper.html}
}

@inproceedings{ye2024atpllava,
  title     = {{ATP-LLaVA}: Adaptive Token Pruning for Large Vision Language Models},
  author    = {Ye, Xubing and Gan, Yukang and Ge, Yixiao and Zhang, Xiao-Ping and Tang, Yansong},
  booktitle = {Proceedings of the IEEE/CVF Conference on Computer Vision and Pattern Recognition},
  pages     = {24972--24982},
  year      = {2025},
  doi       = {10.1109/CVPR52734.2025.02325},
  url       = {https://openaccess.thecvf.com/content/CVPR2025/html/Ye_ATP-LLaVA_Adaptive_Token_Pruning_for_Large_Vision_Language_Models_CVPR_2025_paper.html}
}

@article{takezoe2026learnpruner,
  title         = {{LearnPruner}: Rethinking Attention-Based Token Pruning in Vision Language Models},
  author        = {Takezoe, Rinyoichi and Li, Yaqian and Bo, Zihao and Hou, Anzhou and Guang, Mo and Long, Kaiwen},
  journal       = {arXiv preprint arXiv:2604.23950},
  year          = {2026},
  eprint        = {2604.23950},
  archivePrefix = {arXiv},
  url           = {https://arxiv.org/abs/2604.23950}
}

@article{tschannen2025siglip2,
  title         = {{SigLIP 2}: Multilingual Vision-Language Encoders with Improved Semantic Understanding, Localization, and Dense Features},
  author        = {Tschannen, Michael and Gritsenko, Alexey and Wang, Xiao and Naeem, Muhammad Ferjad and Alabdulmohsin, Ibrahim and Parthasarathy, Nikhil and Evans, Talfan and Beyer, Lucas and Xia, Ye and Mustafa, Basil and H{\'e}naff, Olivier and Harmsen, Jeremiah and Steiner, Andreas and Zhai, Xiaohua},
  journal       = {arXiv preprint arXiv:2502.14786},
  year          = {2025},
  eprint        = {2502.14786},
  archivePrefix = {arXiv},
  url           = {https://arxiv.org/abs/2502.14786}
}

@inproceedings{yu2016refcoco,
  title     = {Modeling Context in Referring Expressions},
  author    = {Yu, Licheng and Poirson, Patrick and Yang, Shan and Berg, Alexander C. and Berg, Tamara L.},
  booktitle = {European Conference on Computer Vision},
  pages     = {69--85},
  year      = {2016},
  doi       = {10.1007/978-3-319-46475-6_5},
  url       = {https://arxiv.org/abs/1608.00272}
}

@inproceedings{goyal2017vqav2,
  title     = {Making the {V} in {VQA} Matter: Elevating the Role of Image Understanding in Visual Question Answering},
  author    = {Goyal, Yash and Khot, Tejas and Summers-Stay, Douglas and Batra, Dhruv and Parikh, Devi},
  booktitle = {Proceedings of the IEEE Conference on Computer Vision and Pattern Recognition},
  pages     = {6904--6913},
  year      = {2017},
  url       = {https://openaccess.thecvf.com/content_cvpr_2017/html/Goyal_Making_the_v_CVPR_2017_paper.html}
}

@incollection{simon1971designing,
  author    = {Simon, Herbert A.},
  title     = {Designing Organizations for an Information-Rich World},
  editor    = {Greenberger, Martin},
  booktitle = {Computers, Communications, and the Public Interest},
  publisher = {The Johns Hopkins Press},
  address   = {Baltimore, MD},
  year      = {1971},
  pages     = {37--72}
}

@misc{mao2025dove,
  author = {Mao, Lingjun and Corona, Rodolfo and Liang, Xin and Yan, Wenhao and Tang, Zineng},
  title = {Images are Worth Variable Length of Representations},
  year = {2025},
  eprint = {2506.03643},
  archivePrefix = {arXiv},
  primaryClass = {cs.CV},
  url = {https://arxiv.org/abs/2506.03643},
}

@misc{duggal2025karl,
  author = {Duggal, Shivam and Byun, Sanghyun and Freeman, William T. and Torralba, Antonio and Isola, Phillip},
  title = {Single-pass Adaptive Image Tokenization for Minimum Program Search},
  year = {2025},
  eprint = {2507.07995},
  archivePrefix = {arXiv},
  primaryClass = {cs.CV},
  url = {https://arxiv.org/abs/2507.07995},
}

@misc{yan2024elastictok,
  author = {Yan, Wilson and Mnih, Volodymyr and Faust, Aleksandra and Zaharia, Matei and Abbeel, Pieter and Liu, Hao},
  title = {{ElasticTok}: Adaptive Tokenization for Image and Video},
  year = {2024},
  eprint = {2410.08368},
  archivePrefix = {arXiv},
  primaryClass = {cs.CV},
  url = {https://arxiv.org/abs/2410.08368},
}

@inproceedings{bachmann2025flextok,
  author = {Bachmann, Roman and Allardice, Jesse and Mizrahi, David and Fini, Enrico and Kar, O{\u g}uzhan Fatih and Amirloo, Elmira and El-Nouby, Alaaeldin and Zamir, Amir and Dehghan, Afshin},
  title = {{FlexTok}: Resampling Images into {1D} Token Sequences of Flexible Length},
  booktitle = {Proceedings of the 42nd International Conference on Machine Learning},
  series = {Proceedings of Machine Learning Research},
  volume = {267},
  pages = {2241--2292},
  year = {2025},
  publisher = {PMLR},
  url = {https://proceedings.mlr.press/v267/bachmann25a.html},
}

@misc{chen2026stat,
  author = {Chen, Zeyuan and Zhang, Kai and Tu, Zhuowen and Xiong, Yuanjun},
  title = {Soft Tail-dropping for Adaptive Visual Tokenization},
  year = {2026},
  eprint = {2601.14246},
  archivePrefix = {arXiv},
  primaryClass = {cs.CV},
  url = {https://arxiv.org/abs/2601.14246},
}

@misc{lu2026adatok,
  author = {Lu, Xiaocheng and Chen, Yuxi and Zhang, Jie and Liu, Jian and Guo, Jingcai and Zhu, Fangqi and Han, Tao and Guo, Song},
  title = {{AdaTok}: Self-Budgeting Image Tokenization with Quality-Preserving Dynamic Tokens},
  year = {2026},
  eprint = {2606.07185},
  archivePrefix = {arXiv},
  primaryClass = {cs.CV},
  url = {https://arxiv.org/abs/2606.07185},
}

@inproceedings{chen2024fastv,
  author = {Chen, Liang and Zhao, Haozhe and Liu, Tianyu and Bai, Shuai and Lin, Junyang and Zhou, Chang and Chang, Baobao},
  title = {An Image is Worth 1/2 Tokens After Layer 2: Plug-and-Play Inference Acceleration for Large Vision-Language Models},
  booktitle = {European Conference on Computer Vision},
  year = {2024},
  url = {https://www.ecva.net/papers/eccv_2024/papers_ECCV/html/10478_ECCV_2024_paper.php},
}

@inproceedings{zhang2025sparsevlm,
  author = {Zhang, Yuan and Fan, Chun-Kai and Ma, Junpeng and Zheng, Wenzhao and Huang, Tao and Cheng, Kuan and Gudovskiy, Denis A and Okuno, Tomoyuki and Nakata, Yohei and Keutzer, Kurt and Zhang, Shanghang},
  title = {{SparseVLM}: Visual Token Sparsification for Efficient Vision-Language Model Inference},
  booktitle = {Proceedings of the 42nd International Conference on Machine Learning},
  series = {Proceedings of Machine Learning Research},
  volume = {267},
  pages = {74840--74857},
  year = {2025},
  publisher = {PMLR},
  url = {https://proceedings.mlr.press/v267/zhang25s.html},
}

@inproceedings{yang2025visionzip,
  author = {Yang, Senqiao and Chen, Yukang and Tian, Zhuotao and Wang, Chengyao and Li, Jingyao and Yu, Bei and Jia, Jiaya},
  title = {{VisionZip}: Longer is Better but Not Necessary in Vision Language Models},
  booktitle = {Proceedings of the IEEE/CVF Conference on Computer Vision and Pattern Recognition},
  year = {2025},
  pages = {19792--19802},
  url = {https://openaccess.thecvf.com/content/CVPR2025/html/Yang_VisionZip_Longer_is_Better_but_Not_Necessary_in_Vision_Language_CVPR_2025_paper.html},
}

@inproceedings{zhang2025vispruner,
  author = {Zhang, Qizhe and Cheng, Aosong and Lu, Ming and Zhang, Renrui and Zhuo, Zhiyong and Cao, Jiajun and Guo, Shaobo and She, Qi and Zhang, Shanghang},
  title = {Beyond Text-Visual Attention: Exploiting Visual Cues for Effective Token Pruning in {VLMs}},
  booktitle = {Proceedings of the IEEE/CVF International Conference on Computer Vision},
  year = {2025},
  pages = {20857--20867},
  url = {https://openaccess.thecvf.com/content/ICCV2025/html/Zhang_Beyond_Text-Visual_Attention_Exploiting_Visual_Cues_for_Effective_Token_Pruning_ICCV_2025_paper.html},
}

@inproceedings{shang2025prumerge,
  author = {Shang, Yuzhang and Cai, Mu and Xu, Bingxin and Lee, Yong Jae and Yan, Yan},
  title = {{LLaVA-PruMerge}: Adaptive Token Reduction for Efficient Large Multimodal Models},
  booktitle = {Proceedings of the IEEE/CVF International Conference on Computer Vision},
  year = {2025},
  pages = {22857--22867},
  url = {https://openaccess.thecvf.com/content/ICCV2025/html/Shang_LLaVA-PruMerge_Adaptive_Token_Reduction_for_Efficient_Large_Multimodal_Models_ICCV_2025_paper.html},
}
